\documentclass[times,final]{elsarticle}
\usepackage{framed,multirow}

\usepackage{amssymb}
\usepackage{amsfonts}
\usepackage{appendix}
\usepackage{amsmath}
\usepackage{algorithm}
\usepackage{algpseudocode}
\usepackage{bm, bbm}
\usepackage{booktabs}
\usepackage{color}
\usepackage{comment}
\usepackage[T1]{fontenc}    % use 8-bit T1 fonts
\usepackage{hyperref}
\hypersetup{colorlinks,breaklinks,
            linkcolor=blue,urlcolor=blue,
            anchorcolor=darkblue,citecolor=darkblue}
\usepackage[utf8]{inputenc} % allow utf-8 input
\usepackage{latexsym}
\usepackage{lineno}
\usepackage{nicefrac}       % compact symbols for 1/2, etc.
\usepackage{graphicx}
\usepackage{microtype}      % microtypography
\usepackage{multirow}
\usepackage{subfigure}
\usepackage{epstopdf}
\usepackage{pifont}
\usepackage{pdflscape}
\usepackage{tabularx}
\usepackage{txfonts}
\usepackage{threeparttable}
\usepackage{url}
\usepackage[dvipsnames]{xcolor}

\usepackage[figuresright]{rotating}

\definecolor{newcolor}{rgb}{.8,.349,.1}

\newtheorem{remark}{Remark}

\begin{document}

\begin{frontmatter}

\title{HomPINNs: homotopy physics-informed neural networks for solving the inverse problems of nonlinear differential equations with multiple solutions}

\author[1]{Haoyang Zheng}\ead{zheng528@purdue.edu}
\author[2]{Yao Huang}
\author[3]{Ziyang Huang}
\author[4]{Wenrui Hao}
\author[1,5]{Guang Lin\corref{cor1}}
\cortext[cor1]{Corresponding author.}
\ead{guanglin@purdue.edu}
\address[1]{School of Mechanical Engineering, Purdue University, West Lafayette, IN 47907, USA}
\address[2]{School of Mathematics and Statistics, Xi'an Jiaotong University, Xi'an, Shaanxi 710049, China}
\address[3]{Department of Mechanical Engineering, University of Michigan, Ann Arbor, Michigan 48109, USA}
\address[4]{Department of Mathematics, Pennsylvania State University, University Park, PA 16802, USA}
\address[5]{Department of Mathematics, Purdue University, West Lafayette, IN 47907, USA}

\begin{abstract}
Due to the complex behavior arising from non-uniqueness, symmetry, and bifurcations in the solution space, solving inverse problems of nonlinear differential equations (DEs) with multiple solutions is a challenging task. To address this, we propose homotopy physics-informed neural networks (HomPINNs), a novel framework that leverages homotopy continuation and neural networks (NNs) to solve inverse problems. The proposed framework begins with the use of NNs to simultaneously approximate {unlabeled} observations {across diverse solutions} while adhering to DE constraints. Through homotopy continuation, the proposed method solves the inverse problem by tracing the observations and identifying multiple solutions. The experiments involve testing the performance of the proposed method on one-dimensional DEs and applying it to solve a two-dimensional Gray-Scott simulation. Our findings demonstrate that the proposed method is scalable and adaptable, providing an effective solution for solving DEs with multiple solutions and unknown parameters. Moreover, it has significant potential for various applications in scientific computing, such as modeling complex systems and solving inverse problems in physics, chemistry, biology, etc.
\end{abstract}

\end{frontmatter}

\section{Introduction}

Inverse problems are popular in various fields of computational physics, such as fluid dynamics \cite{iglesias2013ensemble}, electromagnetics \cite{sarvas1987basic}, and geophysics \cite{sambridge2002monte}. These problems often involve recovering a set of parameters given certain observations or measurements. Traditional techniques for solving inverse problems include optimization-based methods, such as adjoint methods \cite{givoli2021tutorial}, Bayesian inversion \cite{bui2013computational}, and sampling-based methods \cite{christen2005markov}, etc. While the above-mentioned approaches have proven successful in specific research areas, they come with several obstacles, such as high computational costs, a need for significant domain knowledge, or issues with ill-posed problems.

In the last few years, deep learning techniques, specifically physics-informed neural networks (PINNs) \cite{dissanayake1994neural, raissi2019physics}, have shown potential for solving inverse problems more accurately and efficiently. By integrating prior physical knowledge into neural network (NN) architectures, PINNs enable solving inverse problems with superior generalization ability and a reduced need for the number of observations. The concept of PINNs was first introduced by Raissi \textit{et al.} \cite{raissi2019physics} in a series of papers that demonstrated their potential for solving forward and inverse problems in computational physics. The intuition behind PINNs is the incorporation of prior physical knowledge into the loss function. By minimizing this loss function, the NN learns to approximate the solution to the inverse problem while satisfying the underlying physics. Following the idea of PINNs and with the help of automatic differentiation, a Python library \cite{lu2021deepxde} was developed by Lu \textit{et al.} to solve inverse problems in the area of computational science and engineering.

Since then, PINNs and their variants have been successfully applied to solve inverse problems in many different areas. Building upon PINNs, Pang \textit{et al.} \cite{pang2019fpinns} developed fractional PINNs (fPINNs) for solving space-time fractional advection-diffusion equations from scattered and noisy data. Kharazmi \textit{et al.} \cite{kharazmi2021hp} proposed hp-variational PINNs (hp-VPINNs) with domain decomposition to construct both local and global neural network approximations. Liu \textit{et al.} \cite{yang2021b} introduced Bayesian PINNs (B-PINNs) for solving partial differential equations (PDEs), providing accurate predictions and quantifying aleatoric uncertainty with noisy data. Jagtap \textit{et al.} \cite{jagtap2020conservative} tried to enforce flux continuity and average solution at sub-domain interfaces and proposed conservative PINNs (cPINNs) on discrete domains for nonlinear conservation laws. Jagtap \textit{et al.} \cite{jagtap2021extended} extended a generalized space-time domain decomposition approach, extended PINNs (XPINNs), for solving nonlinear PDEs on complex-geometry domains. Wang \textit{et al.} \cite{wang2021eigenvector} investigate the limitations of PINNs in approximating high-frequency or multi-scale functions, and propose novel architectures using spatiotemporal and multi-scale random Fourier features to improve the robustness and accuracy of PINNs. Wang \textit{et al.} \cite{wang2022and} also investigated the training behavior and failures of PINNs using the Neural Tangent Kernel (NTK) framework, identifying a discrepancy in the convergence rate of different loss components, and proposed a novel gradient descent algorithm that adaptively calibrates the convergence rate. Additionally, PINNs have been implemented in various research domains, including fluid dynamics \cite{raissi2020hidden, jin2021nsfnets}, solid mechanics \cite{haghighat2021physics}, molecular dynamics\cite{patel2022thermodynamically}, quantum chemistry \cite{pfau2020ab}, etc. More papers related to solving inverse problems with the use of PINNs can be found in \cite{lu2021deepxde, karniadakis2021physics}.

Despite the advancements and vast potential of PINNs for solving inverse problems demonstrated in the aforementioned works, they may not be well-suited for addressing differential equations (DEs) with multiple solutions since they rely on gradient descent optimization algorithms that converge to a single local minimum. {This limitation is accentuated when dealing with observations from these solutions that are unlabeled. Although PINNs might accurately approximate mixed observations, the lack of labels and difficulty in identifying observations from multiple solutions can prevent even a well-structured neural network from effectively solving the inverse problem.}

In this work, we consider the combination of PINNs and homotopy continuation methods to address this issue by enabling the exploration of multiple solutions for DEs. Homotopy continuation \cite{li1989cheater} is a numerical technique that traces the solution path of a given problem as the parameter changes from an initial to a final value. Hao \textit{et al.} \cite{hao2013homotopy} utilized the WENO scheme in homotopy continuation to resolve steady-state problems associated with hyperbolic conservation laws. Hao \textit{et al.} \cite{hao2014bootstrapping} presented a homotopy continuation-based method using domain decomposition to solve large polynomial systems resulting from discretizing nonlinear algebraic DEs. Hao \textit{et al.} \cite{hao2020adaptive} proposed an adaptive step-size homotopy tracking method for computing bifurcation points in nonlinear systems. Hao \textit{et al.} \cite{hao2020homotopy} developed a novel approach to compute multiple solutions of nonlinear DEs by adaptively constructing a spectral approximation space using a greedy algorithm. These homotopy methods have been successfully applied to solving pattern formation problems arising from math biology \cite{hao2020spatial}.

In this context, we proposed homotopy physics-informed neural networks (HomPINNs), where the homotopy continuation method is incorporated into the PINNs to find multiple solutions by gradually transforming the problem from a simple problem with given conditions to the target deformed problem of interest. The transformation process is accomplished by adjusting the scale of two constraints in the loss function during NN training: one term aims to minimize the distance between the NN approximation and the given observations, while the other term enforces the approximation's conformity with the governing equations. We first ensure that the NN approximation satisfies both two constraints. As training proceeds, we adjust the magnitude of a homotopy tracking parameter to gradually focus more on the first constraint. As the homotopy tracking parameter changes, both DEs with multiple solutions and the inverse problems can be solved. This strategy makes it possible to explore the solution space more effectively and identify various local minima corresponding to different solutions of the DEs.

In contrast to traditional methods used for solving inverse problems of DEs with multiple solutions, our proposed method is advantageous over previous methods in the following aspects:

\begin{enumerate}
  \item {The proposed method can efficiently identify unlabeled observations from multiple solutions.}
  \item The proposed method provides both adaptable and flexible solutions for inverse problems.
  \item The proposed method is ideal for inverse problems that are high-dimensional or computationally intense.
  \item The proposed method is able to solve inverse problems even when only a limited number of DE solutions are available in the observations.
\end{enumerate}

The remainder of this paper is structured as follows: The problem description of the inverse problems and the proposed HomPINN method are presented in Section \ref{method}, along with a comprehensive explanation of the construction details. The effectiveness of our proposed HomPINNs is demonstrated through several numerical examples in Section \ref{sec_test}. The concluding remarks and discussion of our proposed method are presented in Section \ref{sec_discuss}.

\section{The Proposed Framework}\label{method}

We first define the inverse problems associated with DEs. Based on the given observations $\left\{{\boldsymbol x}_i, u_i\right\}_{i=1}^{N_o}$, we detail the technique proposed to identify the correct parameters within DEs whose solution is not unique. {Considering the observations are unlabeled and sourced from various DE solutions, we subsequently elaborate our strategy to identify observations from multiple solutions.} Finally, we delve further into inferring unknown DE parameters from the given observations.

\subsection{Problem description}

We consider the following nonlinear DE:
\begin{equation}\label{equation_pde1}
        \left\{\begin{aligned}
    &\mathcal L u\left({\boldsymbol x}\right)=f\left(u, \boldsymbol x;\boldsymbol \lambda\right),\ \ {\boldsymbol x}\in \Omega,\\
    &\mathcal Bu\left({\boldsymbol x}\right)=b\left({\boldsymbol x}\right),\ \ {\boldsymbol x}\in \partial\Omega.
  \end{aligned}\right.
\end{equation}

Here $\mathcal L$ is a linear differential operator, $f(u, \boldsymbol x;\boldsymbol \lambda)$ is a nonlinear forcing term with an unknown parameter vector $\boldsymbol \lambda$, solution $u$, and $\Omega \in \mathbb R^d$ ($d$ denotes the dimension of the input $\boldsymbol x$) is the domain of interest. $\mathcal B$ is the operator for the boundary condition, and $b\left({\boldsymbol x}\right)$ is the force function at the boundary. Even under the homogeneous boundary condition, i.e., $b(\boldsymbol x)=0$, it is still possible that the solution of (\ref{equation_pde1}) is not unique and that the exact number of solutions is not known. 
Given the observation set $\left\{{\boldsymbol x}_i, u_i\right\}_{i=1}^{N_o}$ ($u_i$ is randomly observed from one of the solutions satisfying (\ref{equation_pde1}) and $N_o$ is the total number of observations), the present study aims to determine the value of $\boldsymbol \lambda$ in (\ref{equation_pde1}).

\subsection{Neural networks to solve differential equations with multiple solutions}\label{section_msnn}

\begin{figure}
  \centering
  \includegraphics[width=6in]{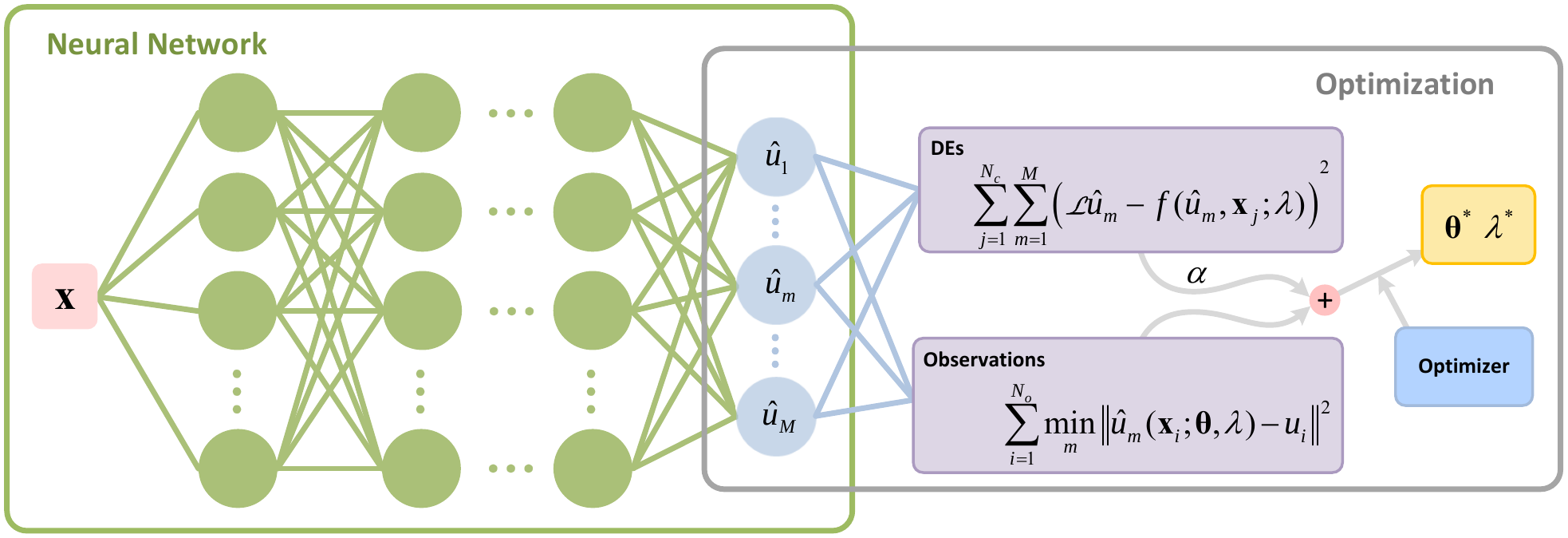}\\
  \caption{The neural network structure of HomPINN for the inverse problems: given the observations $\left\{{\boldsymbol x}_i, u_i\right\}_{i=1}^{N_o}$ and collocation points $\{{\boldsymbol x}_j\}^{N_c}_{j=1}$, HomPINN makes prediction $\boldsymbol{\hat u}$, where $\boldsymbol{\hat u}$ is a concatenation of $\hat u_m$ ($m=1, 2, \cdots, M$). Here $M$ is an estimated total number of solutions of (\ref{equation_pde1}), and $\hat u_m$ (or $\hat u_m({\boldsymbol {x}_i}; \boldsymbol \theta,\boldsymbol \lambda)$) approximates one of the solutions of (\ref{equation_pde1}). Two constraints (purple rectangular boxes) are considered here to constrain the approximations of HomPINNs during the optimization. 
}\label{fig_msnn}
\end{figure}

By leveraging the observations $\left\{{\boldsymbol {x}}_i, u_i\right\}_{i=1}^{N_o}$, collocation points $\{{\boldsymbol {x}}_j\}^{N_c}_{j=1}$ ($N_c$ is the total number of collocation points), and governing equation (\ref{equation_pde1}), we are developing a PINN to find the multiple solutions of (\ref{equation_pde1}). A general framework of the NN can be found in Fig.~\ref{fig_msnn}.
Referring to \cite{raissi2019physics}, we are constructing a NN that can yield an approximation $\boldsymbol{\hat u}({\boldsymbol x_i}; \boldsymbol \theta,\boldsymbol \lambda)$ from input ${\boldsymbol x_i}$, network parameters $\boldsymbol \theta$, and an unknown DE parameter $\boldsymbol \lambda$. Note that $\boldsymbol{\hat u}({\boldsymbol {x}_i}; \boldsymbol \theta,\boldsymbol \lambda)$ is actually a concatenation of $\hat u_{m}({\boldsymbol x}_i;\boldsymbol \theta,\boldsymbol \lambda)$ ($1 \leq m \leq M$), where $\hat u_{m}({\boldsymbol x}_i;\boldsymbol \theta,\boldsymbol \lambda)$ attempts to approximate one of the solutions of (\ref{equation_pde1}), and $M$ is the estimated total number of solutions for (\ref{equation_pde1}). 

\begin{remark}\label{remark_concern}
    One concern is that the estimated $M$ does not match the true quantity of solutions for (\ref{equation_pde1}). We will explore how to accurately identify its correct value in the experimental parts (see \ref{sec_1d_example1} and \ref{sec_seven_sols}).
\end{remark}

For DEs with multiple solutions within the same domain, a follow-up question is to design an appropriate loss function to identify observations into different individual solutions. Merely decreasing the distance (which is measured by traditional loss functions such as mean squared errors, mean absolute errors, etc.) between the NN predictions and observations is insufficient, as the observations from different solutions of (\ref{equation_pde1}) are unlabeled. It could lead to a poor generalization ability of the NNs. A more intelligent strategy would be to design the loss function that can identify the observations and reduce the distances between the NN approximations and observations simultaneously. Specifically, we define the following loss function:

\begin{equation}\label{equation_loss1}
L(\boldsymbol \theta,\boldsymbol \lambda) = \sum^{N_o}_{i=1}\min\limits_{m} \left\|\hat u_{m}({\boldsymbol x}_i;\boldsymbol \theta,\boldsymbol \lambda)-u_i\right \|^2+ \alpha\sum^{N_c}_{j=1}\sum^{M}_{m=1}\big(\mathcal L \hat u_{m}({\boldsymbol x}_j;\boldsymbol \theta,\boldsymbol \lambda)-f(\hat u_{m}({\boldsymbol x}_j;\boldsymbol \theta,\boldsymbol \lambda),{\boldsymbol x}_j;\boldsymbol \lambda)\big)^2.
\end{equation}
Here $\alpha$ is a Lagrange multiplier to balance the weight between the two loss terms. By assuming that the observations are from $M$ or fewer smooth and continuous solutions of (\ref{equation_pde1}), $\hat u_{m}(\cdot)$ is an approximation to one of the solutions. In (\ref{equation_loss1}), the first term captures the proximity of $\hat u_{m}(\cdot)$ to observations at the same ${\boldsymbol x}_i$. In the optimization process, we optimize $\boldsymbol \theta$ (parameters in NNs) and $\boldsymbol \lambda$ (parameters in DEs) to minimize the approximate errors. While the first term helps to approximate observations well, a potential issue is the predictions from $\hat u_{m}(\cdot)$, without incorporating physical knowledge, the predictions given by $\hat u_{m}(\cdot)$ may result in non-smooth and discontinuous approximations. It may also lead to a buildup of errors, which can be hard to correct later. A possible solution is to introduce the second loss term in (\ref{equation_loss1}) which evaluates the residual of (\ref{equation_pde1}) using $\hat u_{m}(\cdot)$ at collocation points and thus avoid discontinuity in the NN approximation by imposing the approximation to conform DE constraints. We will further discuss the strategy to properly weigh the two terms in the next section, such that it can help to solve inverse problems when observations are sampled from multiple DE solutions.

\subsection{Physics-informed neural networks with homotopy process}

\begin{figure}
  \centering
  \includegraphics[width=6.5in]{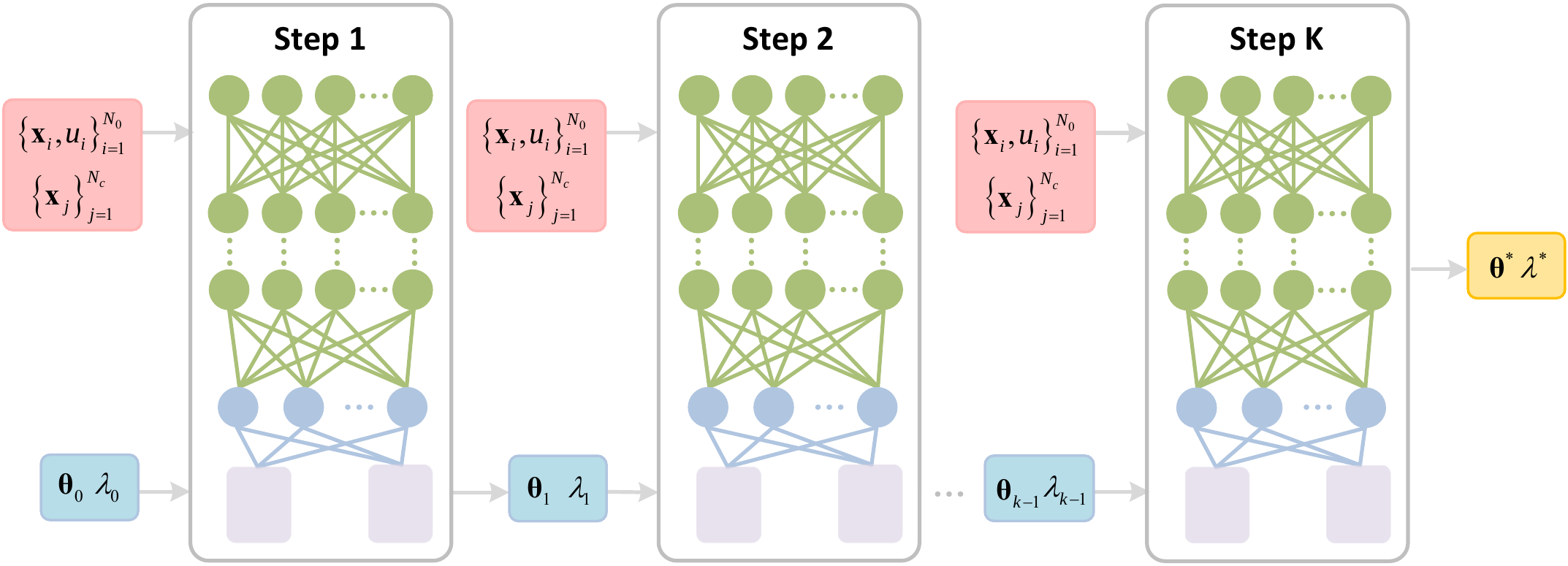}\\
  \caption{Schematic of HomPINNs: a NN at step $k$ ($k=1,2,\cdots, K$) is constructed as Fig.~\ref{fig_msnn}, where the network parameters are initialized by a previous well-trained NN at step $k-1$. Then the observations and collocation points are used to train the NN at each homotopy step. The homotopy process starts from step $1$ and ends in step $K$. The unknown parameter $\boldsymbol \lambda^*$, the optimal parameters $\boldsymbol \theta^*$, and the solutions of (\ref{equation_pde1}) will be obtained finally.}\label{fig_hmsnn}
\end{figure}

%the first  is initialized by {\it HomPINNs} for differential equations (\ref{equation_pde1}) at a given parameter $\lambda$
A well-trained traditional PINN may work for some simple inverse problems, but recovering unknown parameters $\boldsymbol \lambda$ in (\ref{equation_pde1}) with multiple solutions is still challenging. {The difficulty arises from their limitation in identifying unlabeled observations from multiple solutions, leading to a failure in solving the inverse problem.} By leveraging the idea of homotopy continuation \cite{hao2013homotopy} to start with a simpler and related problem with known solutions, track the original problem to the deformed one, and finally identify unknown parameters in (\ref{equation_pde1}), we developed HomPINNs to solve the inverse problem of DEs with multiple solutions. The magnitude of $\alpha$ in the loss function (\ref{equation_loss1}) is utilized to trace from the constraint for satisfying both observations and DEs to the one that focuses more on approximating observations. During the homotopy process, the unknown parameters $\boldsymbol \lambda$ are adjusted to gradually match the observations until the correct $\boldsymbol \lambda$ will make HomPINN approximate the observations perfectly. Fig.~\ref{fig_hmsnn} illustrates a general workflow for executing HomPINNs. HomPINNs feature a shared network structure such that the parameters obtained in the previous homotopy step become the initial guess of the next step. In step $k$ ($1 \leq k \leq K$), the loss function becomes

\begin{equation}\label{equation_loss2}
L_k(\boldsymbol \theta,\boldsymbol \lambda) =\frac{1}{N_o}\sum^{N_o}_{i=1}\min\limits_{m} \left\|\hat u_{m}({\boldsymbol x}_i;\boldsymbol \theta,\lambda)-u_i\right \|^2+ \frac{\alpha_k}{MN_c}\sum^{N_c}_{j=1}\sum^{M}_{m=1}\big(\mathcal L \hat u_{m}({\boldsymbol x}_j;\boldsymbol \theta,\lambda)-f(\hat u_{m}({\boldsymbol x}_j;\boldsymbol \theta,\lambda),{\boldsymbol x}_j;\boldsymbol \lambda)\big)^2,
\end{equation}
where $\alpha_k$ is a homotopy tracking parameter in step $k$. The value of $\alpha_k$ decreases monotonically, i.e., $\alpha_0>\alpha_1>\cdots>\alpha_k>\cdots>\alpha_{K-1}> \alpha_K$. As $\alpha_k$ decreases, the NN will track more of the given observations, and $\boldsymbol \lambda$ will also be adjusted to better fit the observations. {Intuitively, by assigning a rough estimate for $\lambda$ at the beginning, the second term in the loss function acts as a naive DE solver. With this solver, the neural network starts to approximate the observations and align with the DE constraints to avoid discontinuous approximations. It is worth noting that while a rough $\lambda$ might prevent optimal data approximation, it does allow for some level of observation identification. Our empirical findings in section \ref{sec_1d_example1} confirm that the inverse problem can be addressed by starting with a rough estimate of $\lambda$.}

To facilitate the implementation, we further claim that the tracking $\alpha_k$ in this study is structured to decrease exponentially:
\begin{equation}\label{equation_loss3}
\alpha_k=\alpha_0 r^{k-1}.
\end{equation}
{Here $\alpha_0$ serves as an initial value to start the process, which crucially balances data approximation and compliance with DE constraints during the early training phase.} Additionally, $r$ is the decay rate during the homotopy process {($r\in (0, 1)$), which impacts the reduction rate of the multiplier  $\alpha_0 r^{k-1} $ and affects the balance between DE constraints and data approximation. The gradual reduction in $\alpha_0 r^{k-1}$ mirrors the homotopy continuation approach, tracing the solution path from our initial parameter estimation (a simpler solution) to the desired end state. As $\alpha_k$ decreases, the focus shifts from being primarily DE-oriented to emphasizing closer alignment with the observations. In essence, HomPINN adopts a method of iterative refinement: it starts with an initial estimate for the DE parameter, and as the training progresses, the model increasingly focuses on closely matching the observations while adhering to the DE constraints, leading to an effective solution for the inverse problem.}

The training process of HomPINNs is summarized in Fig.~\ref{fig_hmsnn} and Algorithm \ref{algo1}. With the given observations, collocation points, and parameter initializations, HomPINNs employ $K$ homotopy steps inside for optimization. {The selection of homotopy tracking parameters $\alpha_0$ and $r$ often require empirical fine-tuning based on the training's progression. Specifically, if the $\lambda$ value remains static or if training and testing losses show little variation, it is an indicator that they may require adjustment. Through our experiments, we have found that an initial $\alpha_0$ value of 1.00 and the choice of $r=0.60$ consistently perform well across different scenarios.} Additionally, the present study uses the Adam optimizer \cite{Kingma2014AdamAM} starting with a learning rate of $10^{-3}$ and beta values of 0.9 and 0.999 in each homotopy step. Upon finishing the $K$ homotopy steps, the optimized parameters are produced as the final output.

\begin{algorithm}
\caption{The training process of the proposed HomPINNs.}
\hspace*{\algorithmicindent} \textbf{Input} Observations $\{{\boldsymbol x}_i, u_i\}^{N_o}_{i=1}$.\\
\hspace*{\algorithmicindent} \textbf{Input} Collocations $\{{\boldsymbol x}_j\}^{N_c}_{j=1}$.\\
\hspace*{\algorithmicindent} \textbf{Initialize} DE parameter $\boldsymbol\lambda_0$. \\
\hspace*{\algorithmicindent} \textbf{Initialize} HomPINN parameters $\boldsymbol \theta_0$. \\
\hspace*{\algorithmicindent} \textbf{Initialize} Number of outputs $M$. \\
\hspace*{\algorithmicindent} \textbf{Initialize} Homotopy tracking parameters $\alpha_0$, $r$
\begin{algorithmic}[1]
  \For{$k=1,2,\cdots,K$}
    \State Update homotopy tracking parameter $\alpha_k$ from (\ref{equation_loss3}).
        \State Initialize $\boldsymbol \theta=\boldsymbol \theta_{k-1}$, $\boldsymbol \lambda=\boldsymbol \lambda_{k-1}$,
    \State Optimize parameters $\left[\boldsymbol \theta_{k},\boldsymbol \lambda_{k}\right]=\text{argmin}_{\boldsymbol \theta,\boldsymbol \lambda}L_k(\boldsymbol \theta,\boldsymbol \lambda)$, where $L_k(\boldsymbol \theta,\boldsymbol \lambda)$ is computed from (\ref{equation_loss2}).
  \EndFor
\end{algorithmic}\label{algo1}
\hspace*{\algorithmicindent} \textbf{Output} DE parameters $\boldsymbol\lambda_K$.\\
\hspace*{\algorithmicindent} \textbf{Output} PINN parameters $\boldsymbol \theta_K$.
\end{algorithm}

\begin{remark}\label{mark_hmsnn}
In certain situations, particularly when the number of solutions of (\ref{equation_pde1}) is large or the parameter space of $\boldsymbol\lambda$ is in high-dimension, HomPINNs may output values of $\boldsymbol\lambda^*$ that are not accurate enough. Our strategy is to regard the value of $\boldsymbol \lambda^*$ as the hyper-parameter in the HomPINN for the forward problems \cite{huang2022hompinns} (we will call it forward HomPINNs in the following) and train it to obtain the optimized network parameters $\boldsymbol \theta^*$. Once the training is complete, the optimized parameters $\boldsymbol\lambda^*$ and $\boldsymbol \theta^*$ will initialize the NN for inverse problems ($\boldsymbol \lambda_0=\boldsymbol \lambda^*$ and $\boldsymbol \theta_0=\boldsymbol \theta^*$). Then we can run the proposed HomPINNs again to identify the correct $\boldsymbol \lambda$.
\end{remark}

\section{Numerical Examples}\label{sec_test}
In this section, we conduct numerical experiments to verify the proposed methodology\footnote{Code is available at \href{http://github.com/haoyangzheng1996/hompinn}{github.com/haoyangzheng1996/hompinn}.}. We begin with one-dimensional problems to study the effect of hyper-parameters on the algorithm's performance and to provide general rules of thumb for implementing the method. Building on the above experiment, we proceed to solve a two-dimensional elliptic DE problem. The tests mainly run on a desktop with the following specifications: AMD Ryzen Threadripper PRO 5955WX CPU, RTX 4090 GPU, and 128 GB DDR4 RAM memory. The number of collocation points is chosen such that the model performance is not influenced.

\subsection{1D example with two solutions}\label{sec_1d_example1}

The first one-dimensional DE considered is from \cite{hao2014bootstrapping} and can be expressed as:

\begin{equation}\label{equation_ex1}\centering
  \left\{\begin{aligned}
    &\frac{\partial ^2 u(x)}{\partial x^2}=-\lambda\left(1+ u^4\right),\ \ x\in (0,1)\\
    &{\left.\frac{\partial u(x)}{\partial x}\right |_{x=0}=\left.u(x)\right |_{x=1}=0}.
  \end{aligned}\right.
\end{equation}

With $\lambda = 1.20$, (\ref{equation_ex1}) has two solutions, and 80 observations are randomly collected from these two solutions. The two solutions, denoted as $\{u_1(x_i),u_2(x_i)\}$, are from a specific observation location $x_i$, and the observation $u_i$ corresponds to $x_i$ is randomly selected from one of the solutions. Also, all the observation locations $\{x_i\}_{i=1}^{N_o}$ are randomly selected in the domain of interest. Such a strategy of generating observations is applied in all the tests in the present study.

{The NN used in this experiment is a fully connected network, comprising an input layer with a single neuron, three hidden layers with 30 neurons each, and an output layer with $M$ neurons.} The network is initialized following the strategy in \cite{he2015} to start the training in the first homotopy step. We first discuss how to determine $M$ in HomPINN. {We adopt an iterative procedure, where we start with a conservative estimation for $M$ and gradually increment it. We monitor key performance metrics, such as $\lambda$ value, the training and testing losses, and use these as criteria to decide if further increments in $M$ are necessary. Once the losses stabilize within an acceptable range, the iterative process can be halted, and the current $M$ value can be used for the task. This iterative and automatic procedure helps to avoid underestimating $M$.}

Table \ref{table_ex1} summarizes the results with different $M$, which includes the approximated $\lambda$ values, absolute errors in the approximated $\lambda$ (denoted as $\text{err}(\lambda)$), training losses and testing losses identified by HomPINN after one homotopy process. We observe sudden changes in the identified values of $\lambda$, training loss, and testing loss when $M$ changes from $1$ to $2$. This behavior suggests that the exact number of solutions should not be one. As $M$ increases beyond $2$, $\lambda$ value exhibits stability, which points towards $M=2$ being the optimal selection. It is noteworthy that even when $M$ is overestimated, HomPINN can still identify the $\lambda$ value accurately.

\begin{table}[h!]
  \centering
  \begin{threeparttable}
  \caption{Metrics determined by HomPINN following one homotopy process for various $M$ selections. From the second column to the last one, the table indicates $\lambda$ values after one homotopy process, the absolute errors in the approximated $\lambda$ with respect to the benchmark value of $1.20$, the resultant training losses, and testing losses.}
  \label{table_ex1}
    \begin{tabular}{ccccc}
    \toprule
      $M$  & $\lambda$ value & err$(\lambda)$       & Train loss      & Test loss\\    \midrule
      1    &   1.3258     &   $1.26$E-1 & $2.04$E-2   & $1.18$E-1 \\     \midrule
      2    &   1.2000     &   $3.90$E-6 & $1.89$E-9   & $2.13$E-6\\    \midrule
      3    &   1.2000     &   $4.60$E-6 & $4.63$E-9   & $2.07$E-6\\      \midrule
      4    &   1.2000     &   $1.50$E-6 & $2.98$E-9   & $1.98$E-6 \\    \bottomrule
    \end{tabular}
    \end{threeparttable}
\end{table}

\begin{figure}[htbp]
\centering
\makeatletter\def\@captype{figure}\makeatother
% \subfigure[]{\includegraphics[width=2.5in]{fig/simulation/ex1/ex1_01.pdf}\label{fig_ex1_sample}}
\includegraphics[width=2.5in]{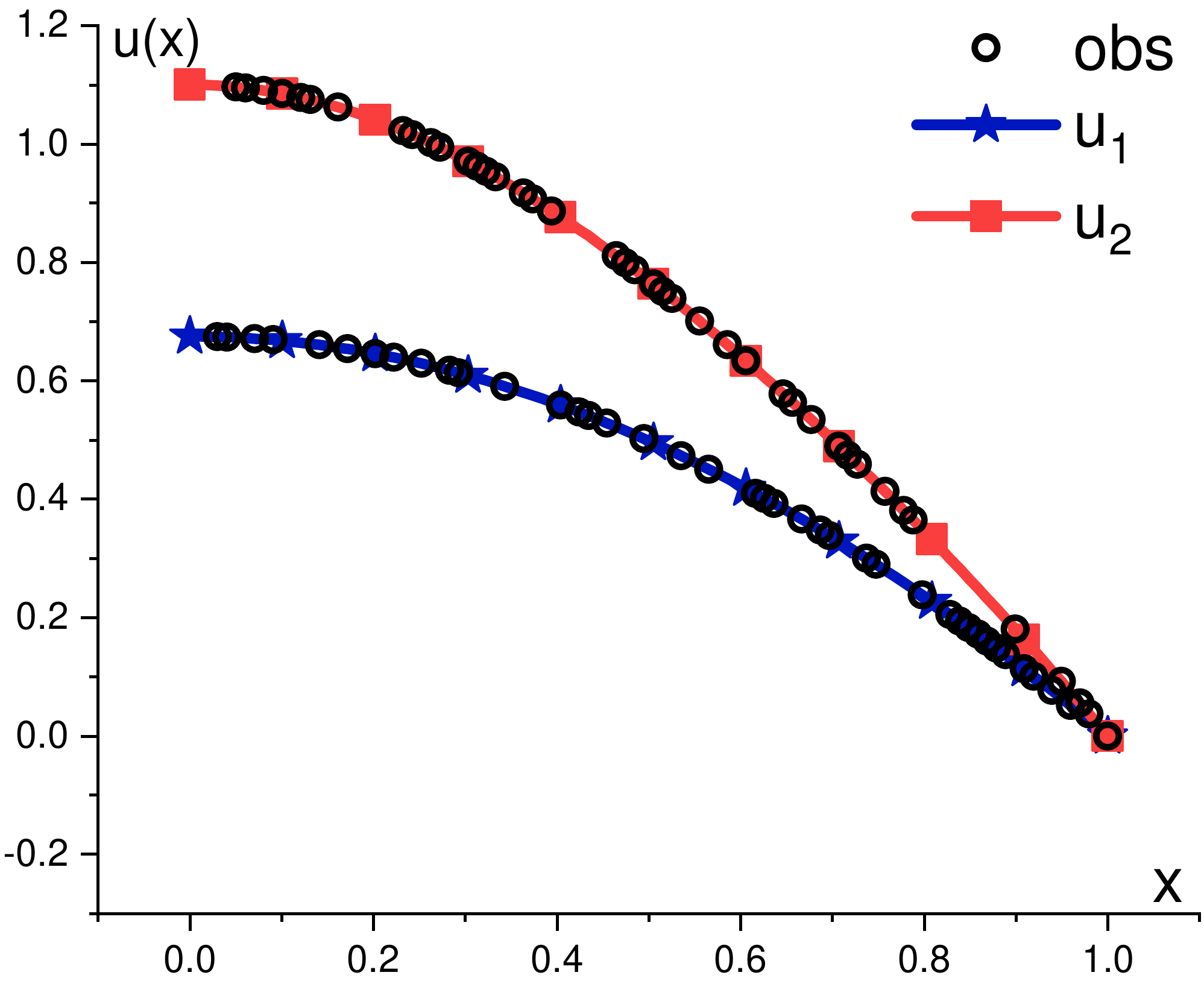}
\caption{Observations sampled from the two solutions of (\ref{equation_ex1}) and approximates made from HomPINNs. Here the x-axis is $x$, and the y-axis is $u$. The black circles are the sampled observations. The red solid line with squares and the blue solid line with stars are the approximated $\hat u_{m}(\cdot)$ from HomPINNs. The correct parameter value $\lambda=1.20$, and with the correct estimate of $M=2$, the identified DE parameter given by HomPINNs $\lambda$ is $1.200$.}\label{fig_ex1_pred}
\end{figure}

\begin{figure}
\centering
% \makeatletter\def\@captype{figure}\makeatother  
\includegraphics[width=6 in]{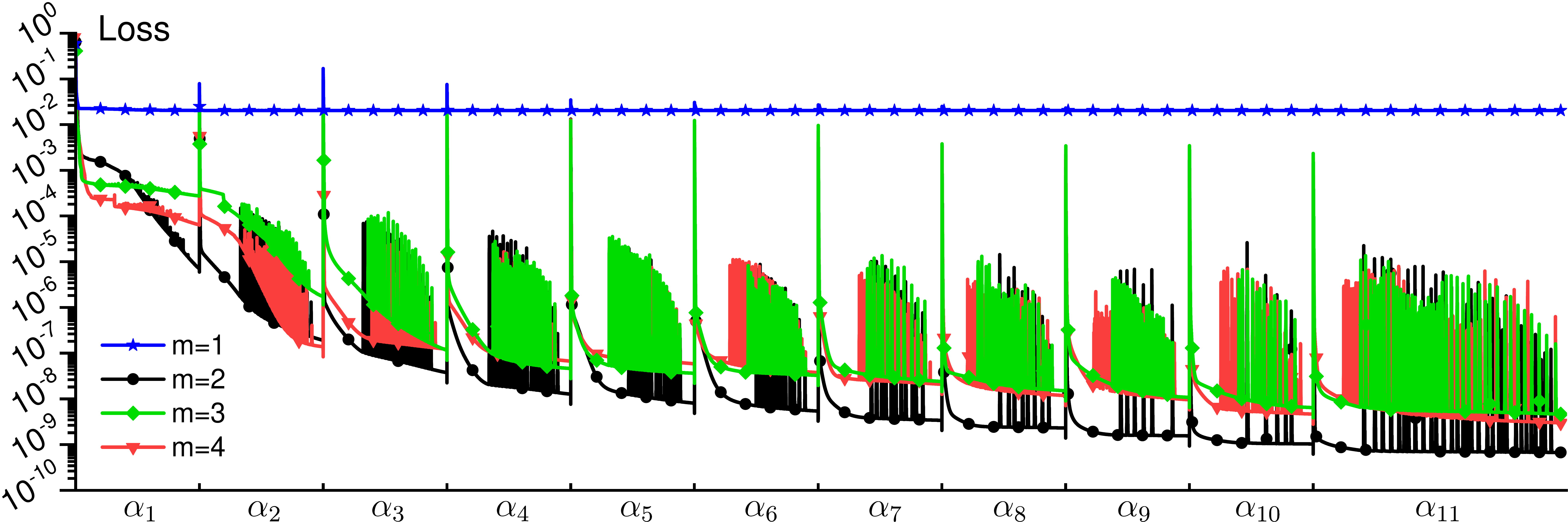}
\caption{Training loss during the homotopy steps with the selection of $M=1,2,3,4$. Here the x-axis indicates the number of homotopy processes performed by HomPINNs, where $\alpha_k$ ($k=1, 2, \cdots, 11$) is the index of each homotopy step. The y-axis is the training loss in the logarithmic scales. }\label{fig_ex1_loss}
\end{figure}

Moreover, the prediction made by the optimized HomPINNs is illustrated in Fig.~\ref{fig_ex1_pred} with $M=2$, along with the observations from the two different solutions of (\ref{equation_ex1}). The change of training loss during the homotopy processes for different $M$ is depicted in Fig.~\ref{fig_ex1_loss}. When using $M=1$, the training loss maintains a value of $2\cdot 10^{-2}$ consistently across the homotopy process. This behavior also illustrates that $M=1$ is not a wise choice for this example. On the other hand, when $M$ is set as 2, 3, or 4, the training losses of HomPINNs decrease as the homotopy steps proceed and converge at a sub-linear rate.

Our next task is to examine the effects of the number of observations ($N_o$). Based on the prior results, we use $M=2$ and $N_o$ values of 80, 60, 40, and 20. The results are shown in Table \ref{table_ex1_2}, which indicates that a smaller amount of observations produces a larger error in the training and testing losses, yet they still remain comparable. These results indicate that the proposed method can still be effective even with fewer observations. This property is favorable when computing resources are limited, but the results suggested that increasing the number of observations helps to derive the HomPINN prediction with high accuracy.

\begin{table}[h!]
  \centering
  \begin{threeparttable}
  \caption{Metrics determined by HomPINN following one homotopy process for various $M$ selections. From the second column to the last one, the table indicates $\lambda$ values after one homotopy process, the absolute errors in the approximated $\lambda$ with respect to the benchmark value of $1.20$, the resultant training losses, and testing losses.}\label{table_ex1_2}
    \begin{tabular}{ccccc}
    \toprule
    $N_o$& $\lambda$ value & err$(\lambda)$ & Train loss& Test loss\\    \midrule
      20    &   1.2000  &   $3.94$E-5 & $6.35$E-9     & $9.49$E-6  \\     \midrule
      40    &   1.2000  &   $2.99$E-5 & $5.81$E-9     & $8.60$E-6  \\    \midrule
      60    &   1.2000  &   $1.13$E-5 & $2.26$E-9     & $7.66$E-6  \\      \midrule
      80    &   1.2000  &   $0.39$E-5 & $1.89$E-9     & $2.13$E-6  \\    \bottomrule
    \end{tabular}
    \end{threeparttable}
\end{table}

\begin{figure}[htbp]
% \centering\includegraphics[width=4 in]{./Figure/violin02.pdf}
\centering\includegraphics[width=3.8 in]{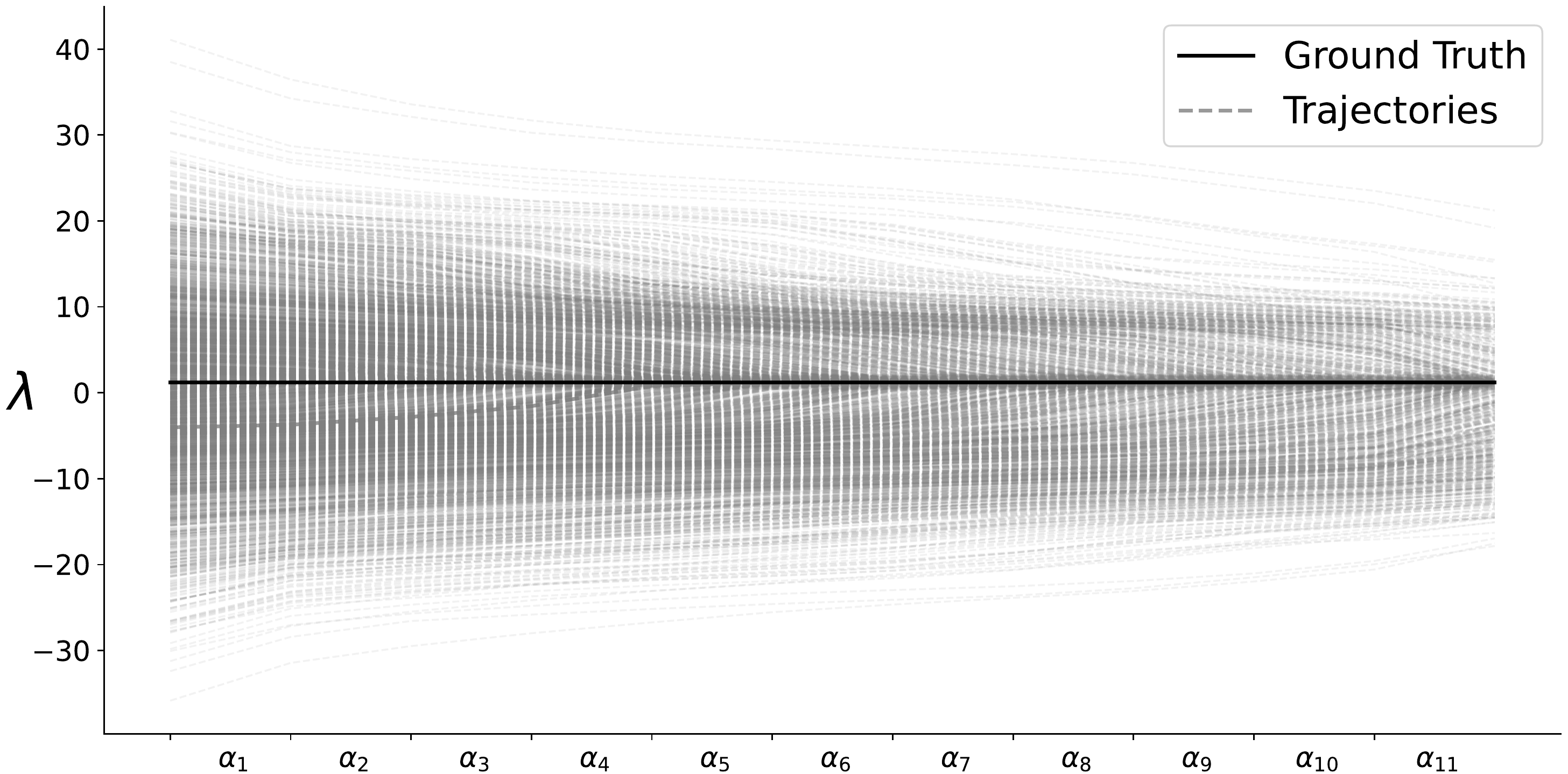}
\centering\includegraphics[width=3.9 in]{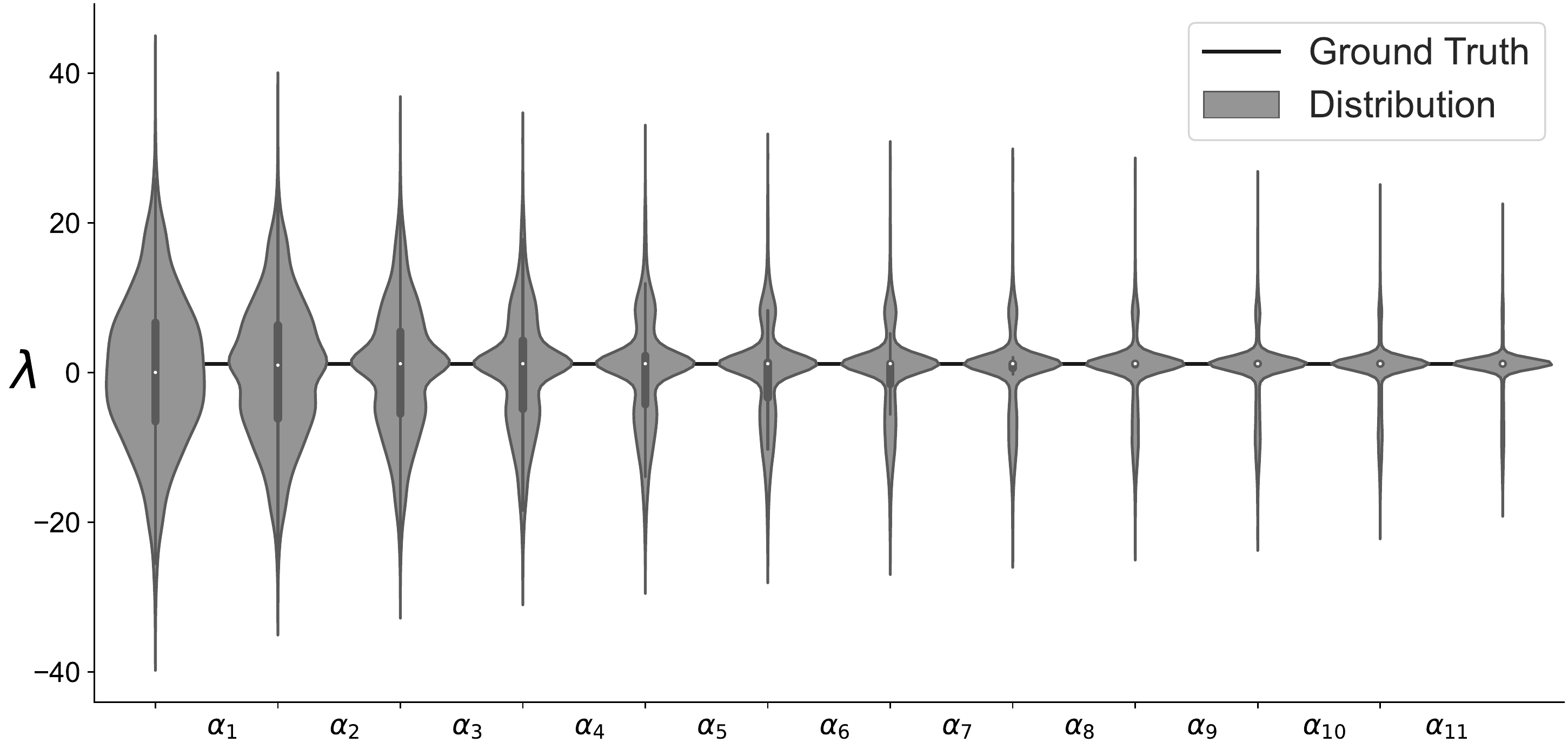}
\caption{{The trajectory of each homotopy training session over its associated number of homotopy steps. The x-axis corresponds to the homotopy step index, while the y-axis represents the values of the unknown DE parameter $\lambda$. \textbf{Left:} Individual trajectories are depicted as gray dashed lines, each of which illustrates the temporal evolution of $\lambda$ during the homotopy training process. \textbf{Right:} For the violin chart, the width of the PDF serves as an indicator of relative frequency for each observed $\lambda$ value. The top and bottom boundaries of the inner box correspond to the third and first quantiles, respectively, and a white circle within the box signifies the median value. Thin black lines extending from the box indicate the lambda's minimum and maximum values. The ground truth: $\lambda=1.20$.}}\label{fig_initialization}
\end{figure}

{We further validate the robustness of the proposed method by solving the inverse problem with diverse initialization of $\lambda$ values. We randomly initialize $\lambda$ from a normal distribution, centered at zero with a standard deviation of ten (Figure \ref{fig_initialization}). Across 3,000 randomly generated instances, a significant portion (about 80\%) converged to the true value of 1.20 after 11 homotopy steps. It is pertinent to note that the remaining instances, even if not converging to the precise value, demonstrated marked improvement from their starting points. This gradual progression toward the correct solution indicates the possibility of refining convergence through minor adjustments to hyper-parameters such as learning rate or the number of homotopy steps. In essence, the robustness of HomPINN against diverse initial conditions, especially when compared with traditional PINNs, underscores its potential to effectively solve complex inverse problems.}

\subsection{1D example with seven solutions}\label{sec_seven_sols}
We delve into a more intricate example, which has 7 solutions \cite{hao2020adaptive}:

\begin{equation}\label{equation_ex2}\centering
  \left\{
  \begin{aligned}
     &\frac{\partial ^2 u(x)}{\partial x^2} =u^4-\lambda u^2,\ \ x\in (0\ ,1),\\
     &\left.\frac{\partial u(x)}{\partial x}\right |_{x=0}=\left.u(x)\right |_{x=1}=0.
  \end{aligned}\right.
\end{equation}

The proposed HomPINN attempts to identify the value of $\lambda=18.00$ given 210 observations ($N_o=210$), where the observations $\{x_i,u_i\}_{i=1}^{210}$ are randomly selected from 7 solutions of (\ref{equation_ex2}). However, we discover that using HomPINN once is unable to provide an accurate enough output of $\lambda$ because the current example is more complex. To alleviate this inaccuracy, we implement the strategy in Remark \ref{mark_hmsnn}. Our test shows that two processes of HomPINN are sufficient in this example.

{To identify the number of solutions, we adopt a similar iterative strategy delineated in Section \ref{sec_1d_example1}, where the results are given in Table \ref{tab_ex2_multiple1}.}  This table  presents the $\lambda$ values, absolute errors in the approximated $\lambda$ (denoted as $\text{err}(\lambda)$), and the training loss identified by HomPINN after two homotopy processes when different $M$ values are considered. Notably, there is a significant decrease in error when transitioning from $M=6$ to $M=7$, suggesting that observations are generated from 7 different solutions governed by (\ref{equation_ex2}). 
\begin{table}[h!]
  \centering
  \begin{threeparttable}
  \caption{Metrics determined by HomPINN following two homotopy processes for various $M$ selections. From the second column to the last column, the table indicates $\lambda$ values post both the first and second homotopy processes, the absolute errors in the approximated $\lambda$ with respect to the benchmark value of $18.00$ after two homotopy processes, and the resultant training losses.}\label{tab_ex2_multiple1}
    \begin{tabular}{ccccc}
    \toprule
    $M$&$\lambda$ (1st process)&$\lambda$ (2nd process)& $\text{err}(\lambda)$& Train loss \\    \midrule
  5    & 17.3050 & 17.2868  &   $7.13$E-1     & $1.51$E-1 \\     \midrule
  6    & 17.4538 & 17.6210  &   $3.79$E-1     & $3.65$E-2 \\    \midrule
  7    & 17.3516 & 17.9982  &   $1.84$E-3     & $2.15$E-5 \\      \midrule
  8    & 17.2918 & 17.9987  &   $1.26$E-3     & $1.79$E-5 \\    \bottomrule
    \end{tabular}
    \end{threeparttable}
\end{table}

\begin{figure}[htbp]
\centering

\subfigure[]{
\includegraphics[width=2.5in]{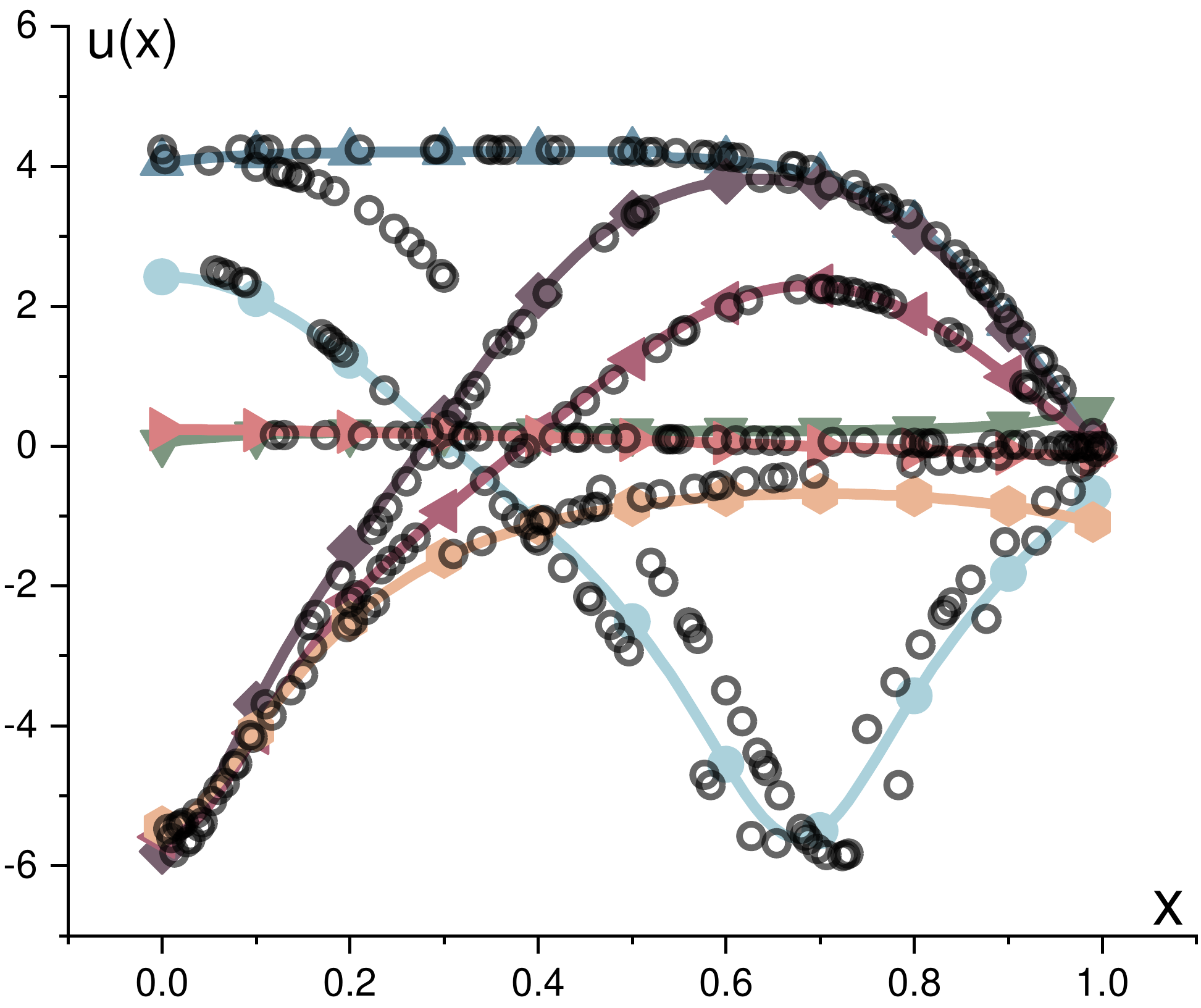}\label{fig_ex2_inverse1}}
\subfigure[]{
\includegraphics[width=2.5in]{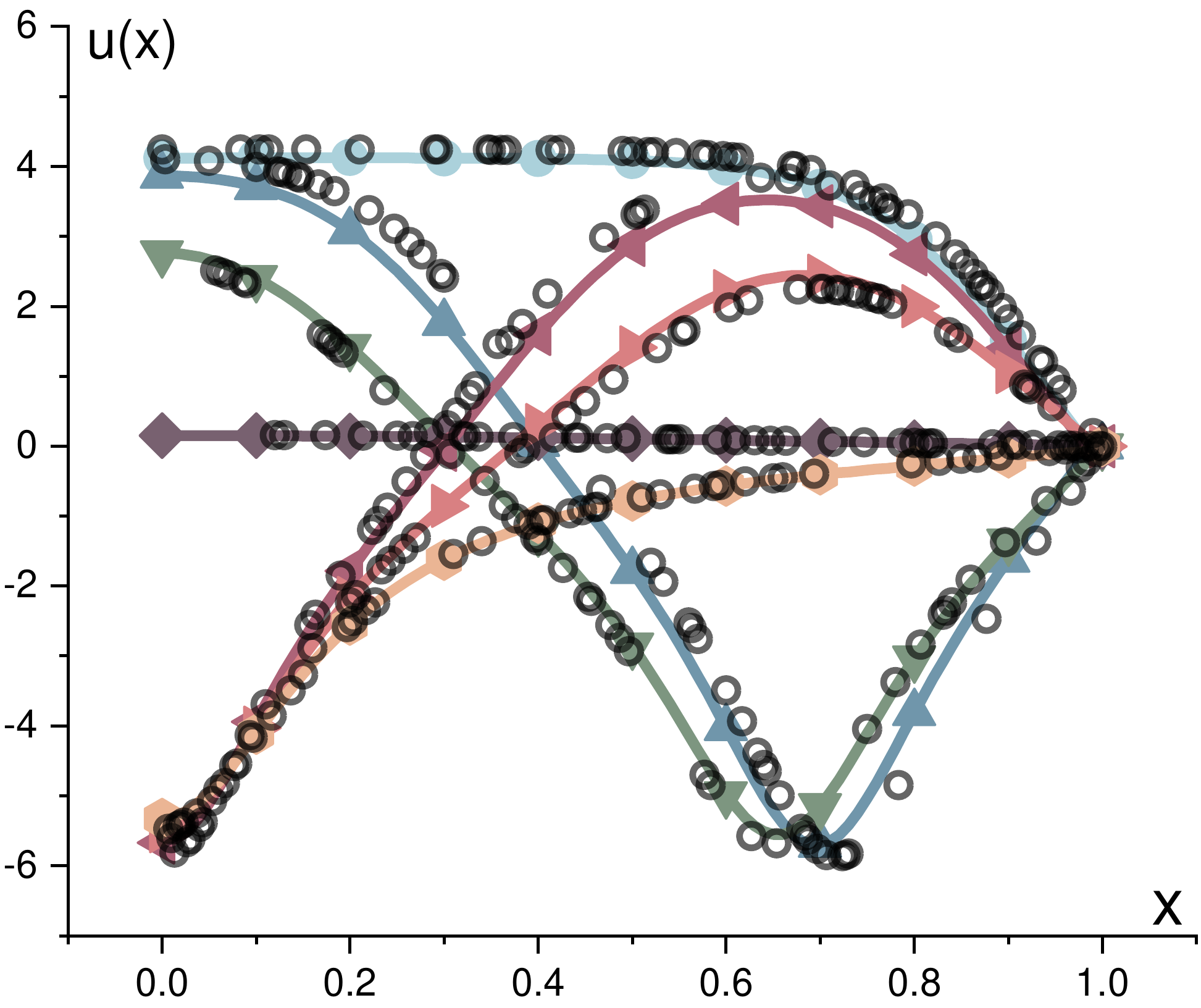}\label{fig_ex2_forward1}}
\subfigure[]{
\includegraphics[width=2.5in]{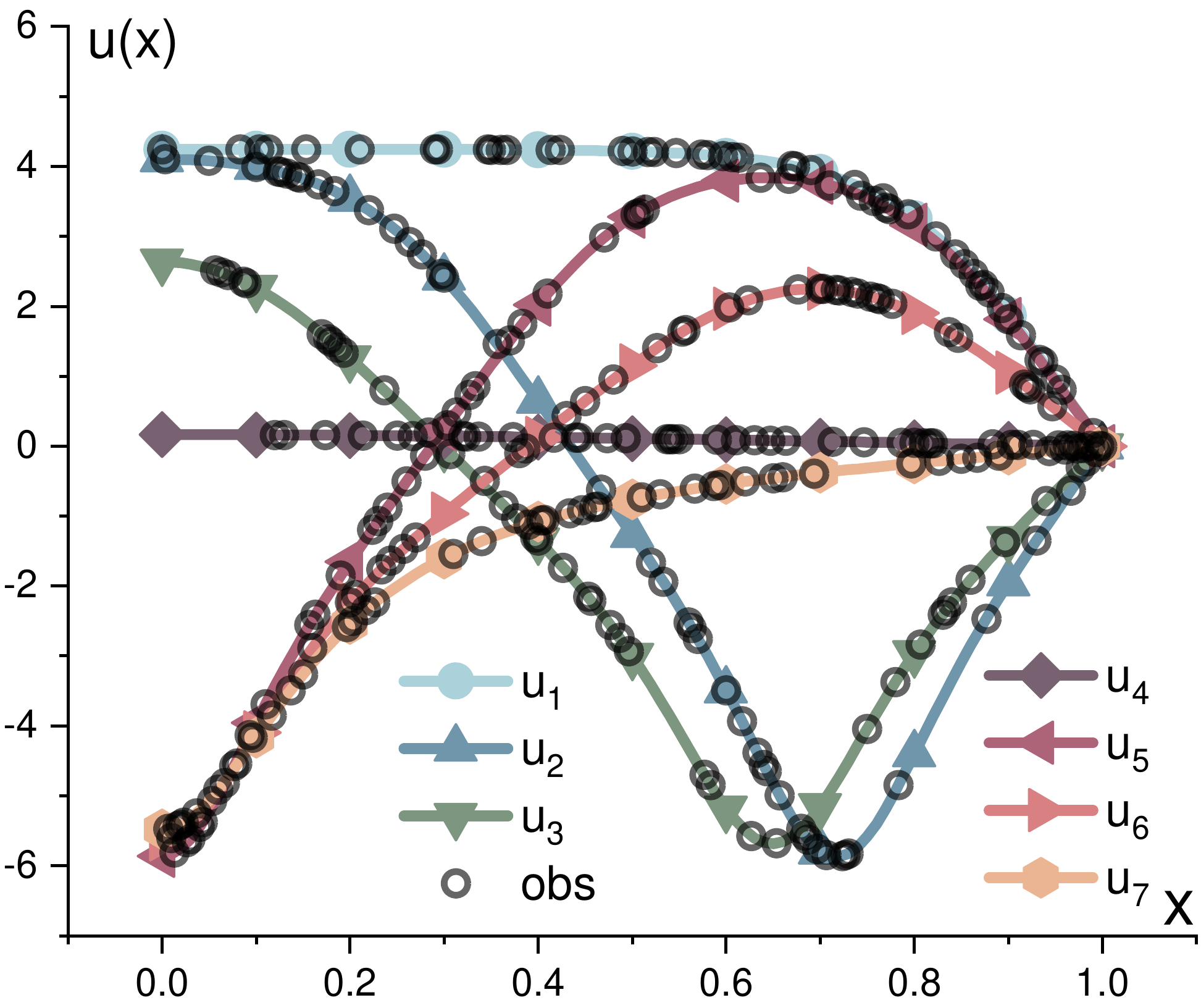}\label{fig_ex2_inverse2}}
\caption{Results from two processes of HomPINN. Here the x-axis is $x$, and the y-axis is $u$. The black circles are observations sampled from the 7 solutions, and seven solid lines represent the approximated $\hat u_{m}(\cdot)$ of (\ref{equation_ex2}). (a): the results of the first homotopy process whose output $\lambda$ is $17.3516$. (b): the results from the forward HomPINNs given DE parameter $\lambda=17.3516$. (c): the results of second homotopy process whose output $\lambda$ is $17.9982$.} \label{fig_ex2_results}
\end{figure}

From Fig.~\ref{fig_ex2_inverse1}, we gather predictions after HomPINN's first complete inverse homotopy process, rendering a $\lambda$ value of 17.3516. Using this $\lambda$ value, we trained the forward HomPINNs, with resulting outputs $\hat u_{m}(\cdot)$ ($m=1, 2,\cdots, 7$) illustrated in Fig.~\ref{fig_ex2_forward1}. Incorporating this $\lambda$ and the NN parameters from the trained forward HomPINNs into HomPINN's second homotopy process yields results in Fig.~\ref{fig_ex2_inverse2}. It is clear from Fig.~\ref{fig_ex2_forward1} that HomPINN, even with an approximate $\lambda$ value from a single homotopy process, approximates $\hat u_{m}(\cdot)$ with heightened accuracy. This implies the NN parameters in the forward HomPINN are approaching their optimal state, refining the accuracy of $\lambda$ in subsequent processes. After determining $\lambda$ via HomPINNs, another forward HomPINN provides solutions for (\ref{equation_ex2}), evident in Fig.~\ref{fig_ex2_inverse2}.

% More insights can be found in Fig.~\ref{fig_ex2_results}, which displays predictions from one complete process of HomPINN, and the corresponding output of $\lambda$ is $17.3516$. Then, we use $\lambda=17.3516$ as an input to train the forward HomPINNs, whose output of $u$ are depicted in Fig.~\ref{fig_ex2_forward1}. Then, we use $\lambda=17.3516$ and the NN parameters of the trained forward HomPINNs to start the second homotopy process of HomPINN, and its results are shown in Fig.~\ref{fig_ex2_inverse2}. As shown in Fig.~\ref{fig_ex2_forward1}, even though running HomPINN the first homotopy process only outputs a value of $\lambda$ that is not very accurate, HomPINN provides a better prediction of $u$ with the given value of $\lambda$ from HomPINN. In other words, the NN parameters ($\boldsymbol \theta$) in the trained forward HomPINN are closer to their global optimum. As a result, using these NN parameters to restart the second homotopy process of HomPINNs helps to find a more accurate value of $\lambda$. Once $\lambda$ is specified from HomPINNs, we use another forward HomPINNs to obtain all the solutions of (\ref{equation_ex2}), as shown in Fig.~\ref{fig_ex2_inverse2}.

% \subsubsection{Observations selected from all solutions}

\begin{figure}
\centering
\makeatletter\def\@captype{figure}\makeatother    
\subfigure{
\includegraphics[width=2.5in]{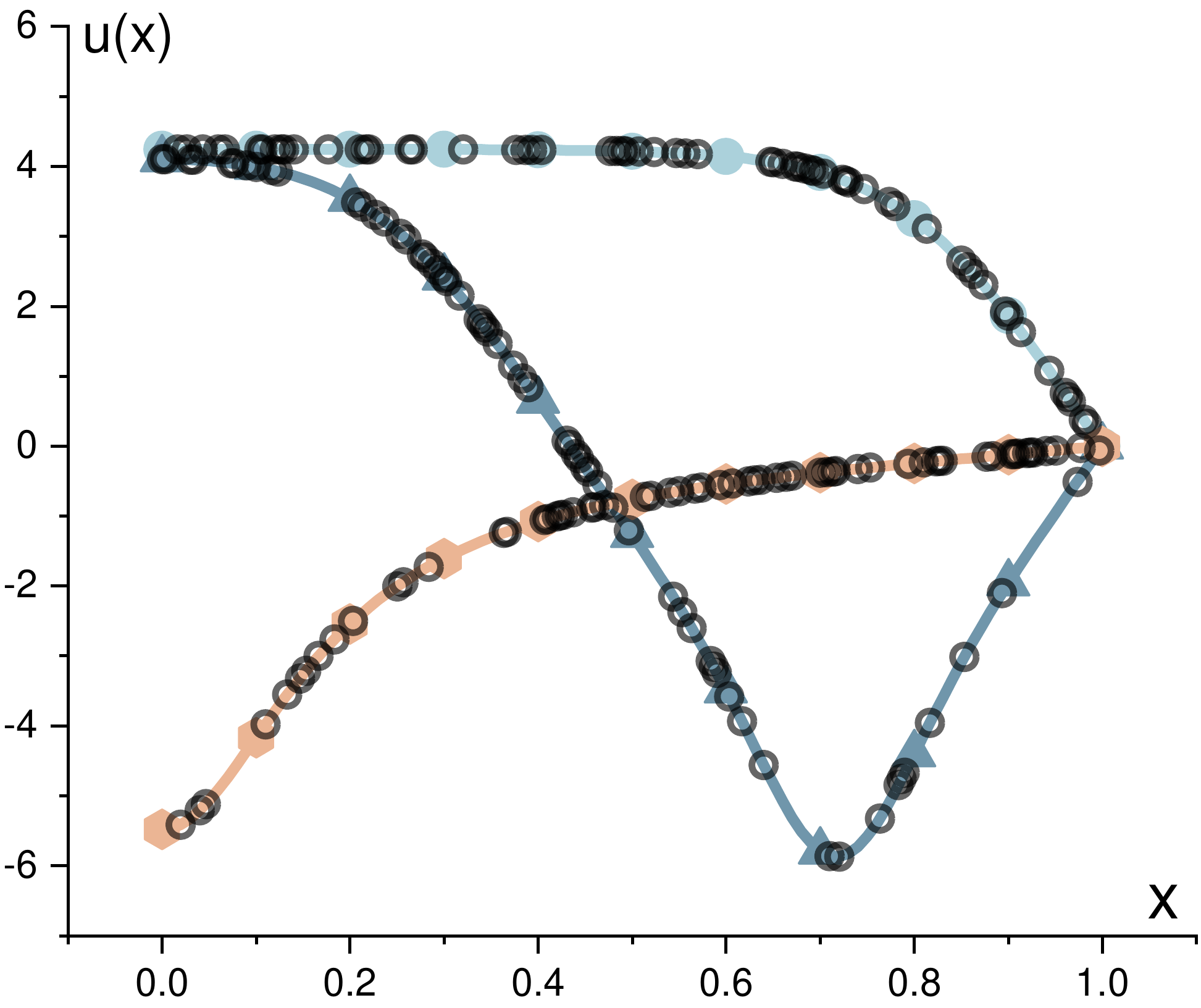}\label{fig_ex2_msnn1}}
\subfigure{
\includegraphics[width=2.5in]{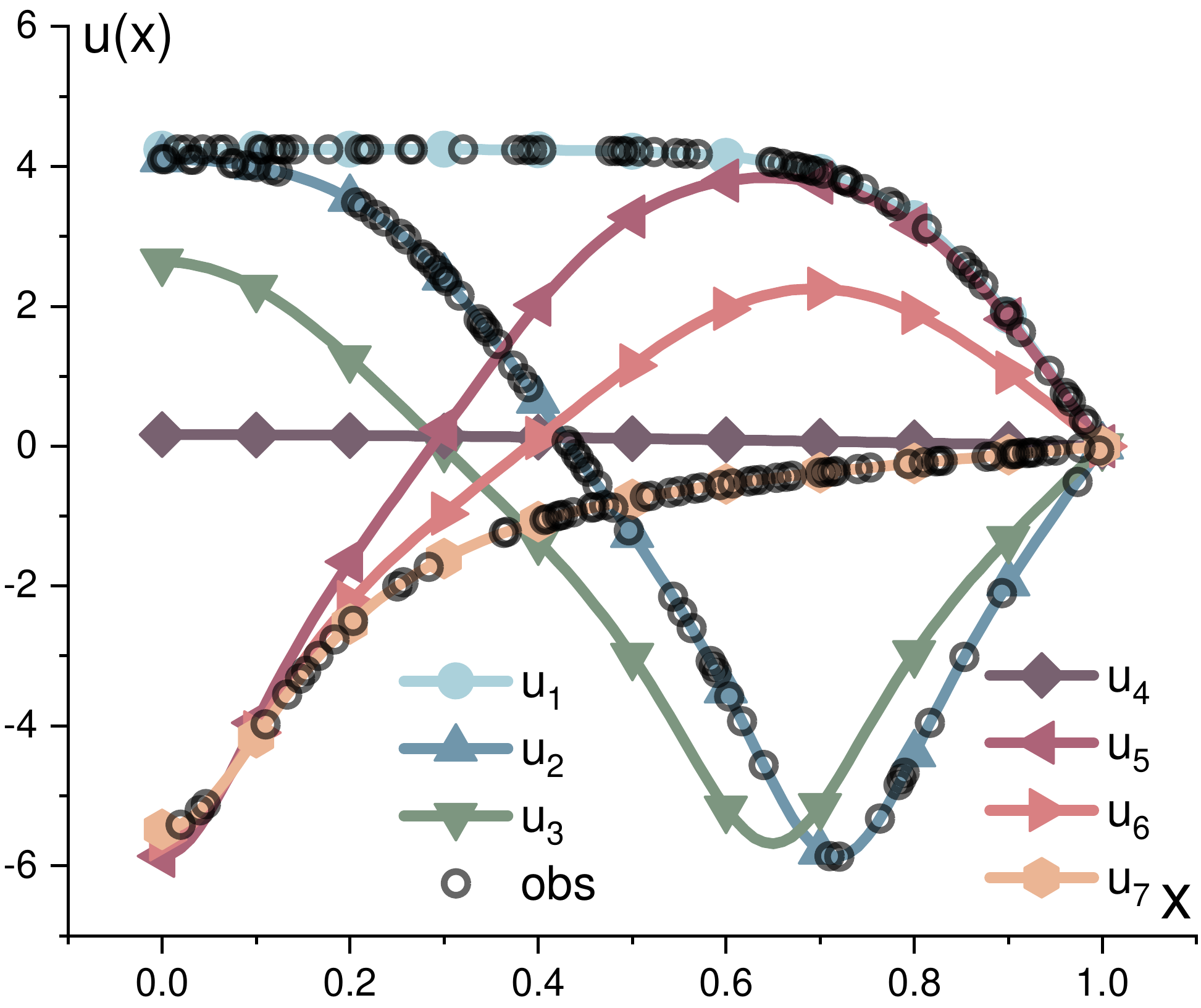}\label{fig_ex2_msnn2}}
\caption{The results after one homotopy process and the predictions made by the forward HomPINNs when only three out of the seven solutions are given. Here x-axis is $x$, and the y-axis is $u$. The black circles are observations sampled from three of the seven solutions, and solid lines represent the approximated $\hat u_{m}(\cdot)$ of (\ref{equation_ex2}). Left:  the trained HomPINNs after one homotopy process. The recovered DE parameter: $\lambda=17.9983$. Right: all the potential solutions of (\ref{equation_ex2}) recovered by the forward HomPINNs given $\lambda=17.9983$.}\label{fig_ex2_pred}
\end{figure}

\begin{table}[h!]
  \centering
  \begin{threeparttable}
  \caption{Metrics determined by HomPINN following one homotopy process for various $M$ selections. From the second column to the last one, the table indicates $\lambda$ values after the first homotopy process, the absolute errors in the approximated $\lambda$ with respect to the true value of $18.00$, and the resultant training losses.}\label{tab_ex2_multiple2}
    \begin{tabular}{cccc}
    \toprule
    $M$& $\lambda$ value & err$(\lambda)$ & Train loss\\     \midrule
  1    & $6.1780$  &   $7.13$E-1     & $8.86$E0  \\     \midrule
  2    & $15.9029$ &   $3.79$E-1     & $1.46$E0 \\    \midrule
  3    & $17.9983$ &   $1.84$E-3     & $5.77$E-5 \\      \midrule
  4    & $17.9989$ &   $1.26$E-3     & $1.62$E-5  \\    \bottomrule
    \end{tabular}
    \end{threeparttable}
\end{table}

It is encouraging to note that with observations from a subset of solutions constrained by DEs, our HomPINNs demonstrate the capability to pinpoint unknown parameters accurately. Notably, for observations $\{x_i,u_i\}^{180}_{i=1}$ from three specific solutions, we achieved an optimal $\lambda=17.9983$ after one homotopy process. We then employed forward HomPINNs, using this $\lambda$, to approximate $\hat u_{m}(\cdot)$ (see Fig.~\ref{fig_ex2_msnn2}). The associated $\lambda$ data and training losses are summarized in Table \ref{tab_ex2_multiple2}. These findings suggest that fewer solution subsets might simplify HomPINN training, particularly in reducing complexities arising from solution overlaps or intersections in high-dimensional spaces.

\begin{figure}
\centering
\makeatletter\def\@captype{figure}\makeatother
\includegraphics[width=6 in]{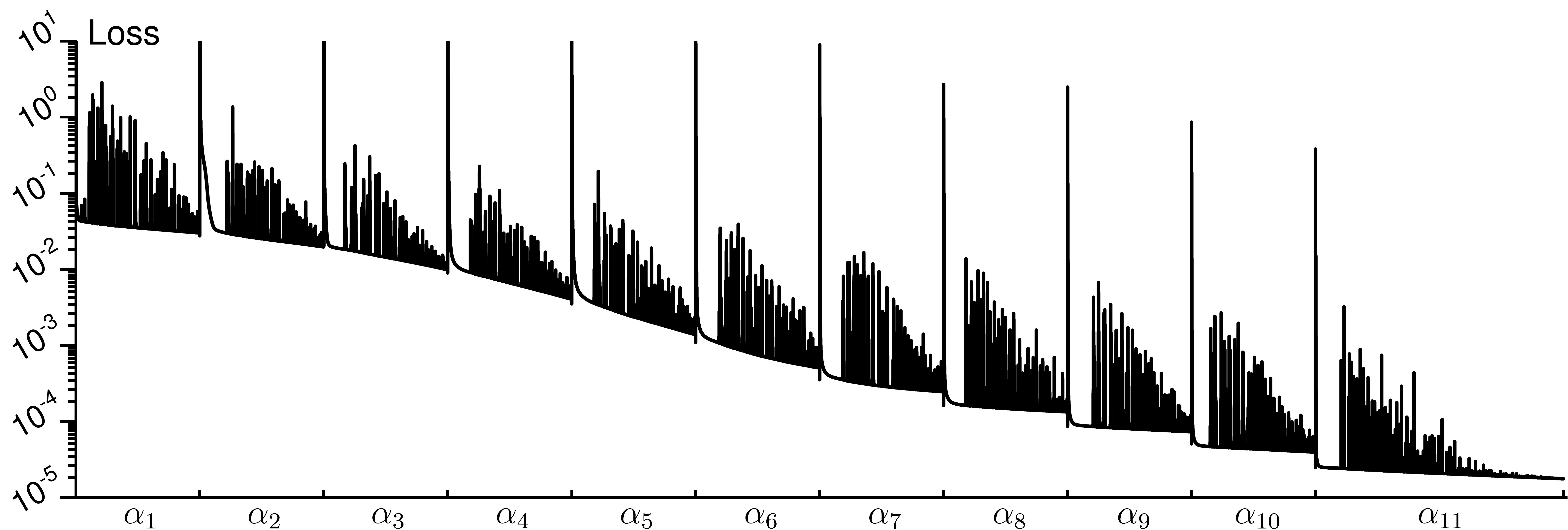}
\caption{The training loss in HomPINN of the example in section \ref{sec_seven_sols}. Here $\alpha_k$ ($k=1,2,\cdots,11$) in the x-axis is the indices of the homotopy step, and the y-axis is the training loss (logarithmic scale). The observations for this training are sampled from three of the seven solutions, and the black solid line is the training loss in HomPINN when $M=3$. The training loss decreases within each homotopy step, and the output value of $\lambda$ is $17.9983$.}\label{fig_ex2_loss}
\end{figure}

\begin{figure*}[hbtp]
\centering
\centering\includegraphics[width=5. in]{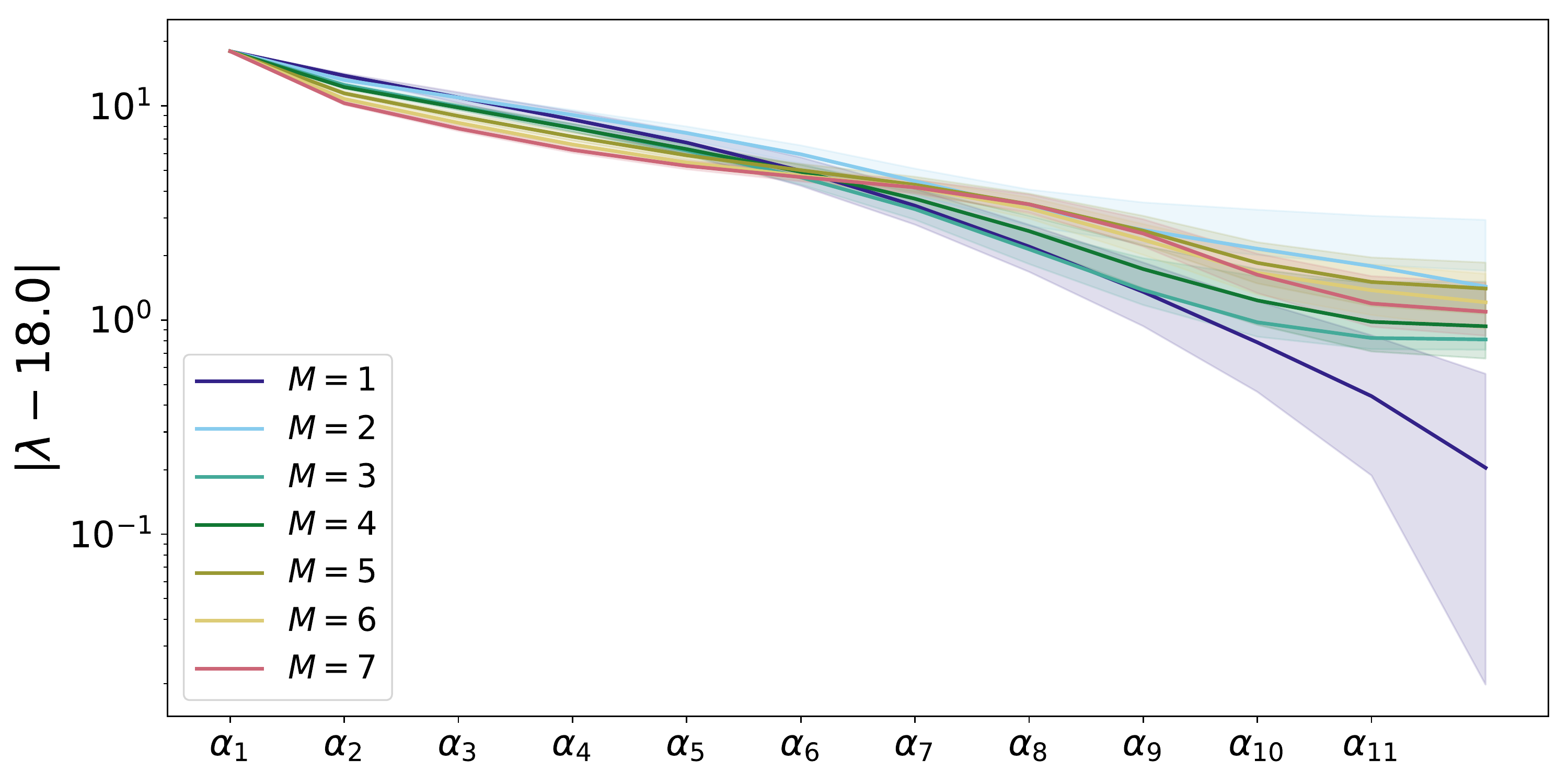}
\caption{{The absolute difference between the derived DE parameter $\lambda$ and the true value of 18, and the corresponding 95\% confidence intervals of the $\lambda$ value as the homotopy steps increase. Each solid line represents the results where observations are sampled from a different number of the seven available solutions, which also corresponds to $M$ selections of HomPINNs.}}\label{fig_mul_sols}
\end{figure*}

{
Building on this, we further evaluated the capability of HomPINN to adeptly solve inverse problems when different numbers of solutions were sampled from the DEs. To achieve this, our experimental design is constructed sequentially, starting with observations randomly and uniformly sampled from any one of the seven available solutions. This sampling process is not only randomized but also repeated in 100 trials to ensure a comprehensive representation of the solution space. Subsequently, each experiment incorporated an additional solution into the same sampling strategy, while solutions were still selected from the full set of seven solutions randomly and uniformly. This gradual expansion continues until the HomPINN is tested against observations drawn simultaneously from all seven solutions. Following each subsequent experiment of 100 trials, we processed the outcomes to derive the $\lambda$ values of each homotopy step and compared it to the true value of 18, thereby quantifying the performance of HomPINN in solving inverse problems under varying degrees of complexity. We then used Bootstrap to get the 95\% confidence intervals and plotted these on a log scale in Fig.~\ref{fig_mul_sols}.
}

{
The study found DE parameter identification to be more straightforward with observations from a single solution  (similar to the standard PINN method). Although integrating more solutions posed increased complexity, our method steadily yielded estimations near the true DE parameter, which validates HomPINN's ability to handle complex scenarios.
}

\subsection{Gray–Scott simulations}

The Gray-Scott model mathematically represents the reaction-diffusion process of two chemical species interacting with each other and their environment. Its origins date back to the 1980s when Gray and Scott \cite{gray1983autocatalytic} first introduced it. Since then, it has been used to simulate a wide range of phenomena, such as pattern formation \cite{hao2020spatial}, chemical waves \cite{farr1992rotating}, and chemical turbulence \cite{mikhailov2006control}. The model describes the changes in concentration of the two species, $A(x, y, t)$ and $S(x, y, t)$. The present study considers the steady state of the Gray-Scott model with the zero-flux boundary condition:
\begin{equation}\label{eq_ex3_steady}
  \left\{\begin{aligned}
    &D_A \Delta A+S A^2-(\mu+\rho) A =0, \\
    &D_S \Delta S-S A^2+\rho(1-S) =0, \\
    &\left.\displaystyle\mathbf{n} \cdot \nabla A \right|_{\partial \Omega}=\left.\displaystyle\mathbf{n} \cdot \nabla S\right|_{\partial \Omega} =0.
\end{aligned}\right.
\end{equation}
Here $\mathbf{n}$ denotes the outward normal at the domain boundary, the two species have diffusion coefficients represented by $D_A$ and $D_S$, respectively. Parameter $\rho$ denotes the constant supply of species $S$ into the system, and $\mu$ represents the rate of decay of species $A$.

To generate observations, we use $D_A=2.5 \times 10^{-4}$, $D_S=5 \times 10^{-4}$, $\rho=4.0 \times 10^{-2}$, and $\mu=6.5\times 10^{-2}$. The observations do not contain all the solutions of (\ref{eq_ex3_steady}) with the parameters chosen. Our goal is to utilize HomPINNs to identify the parameter value and retrieve all the solutions of (\ref{eq_ex3_steady}) given the parameters from HomPINNs. To tackle this complicated task, we first employed trial and error to determine the number of observations ($N_o$) and NN structures (the number of neurons in each layer and the number of layers) that are most effective for HomPINNs.

\begin{figure}[htbp]
\includegraphics[width=1.\columnwidth]{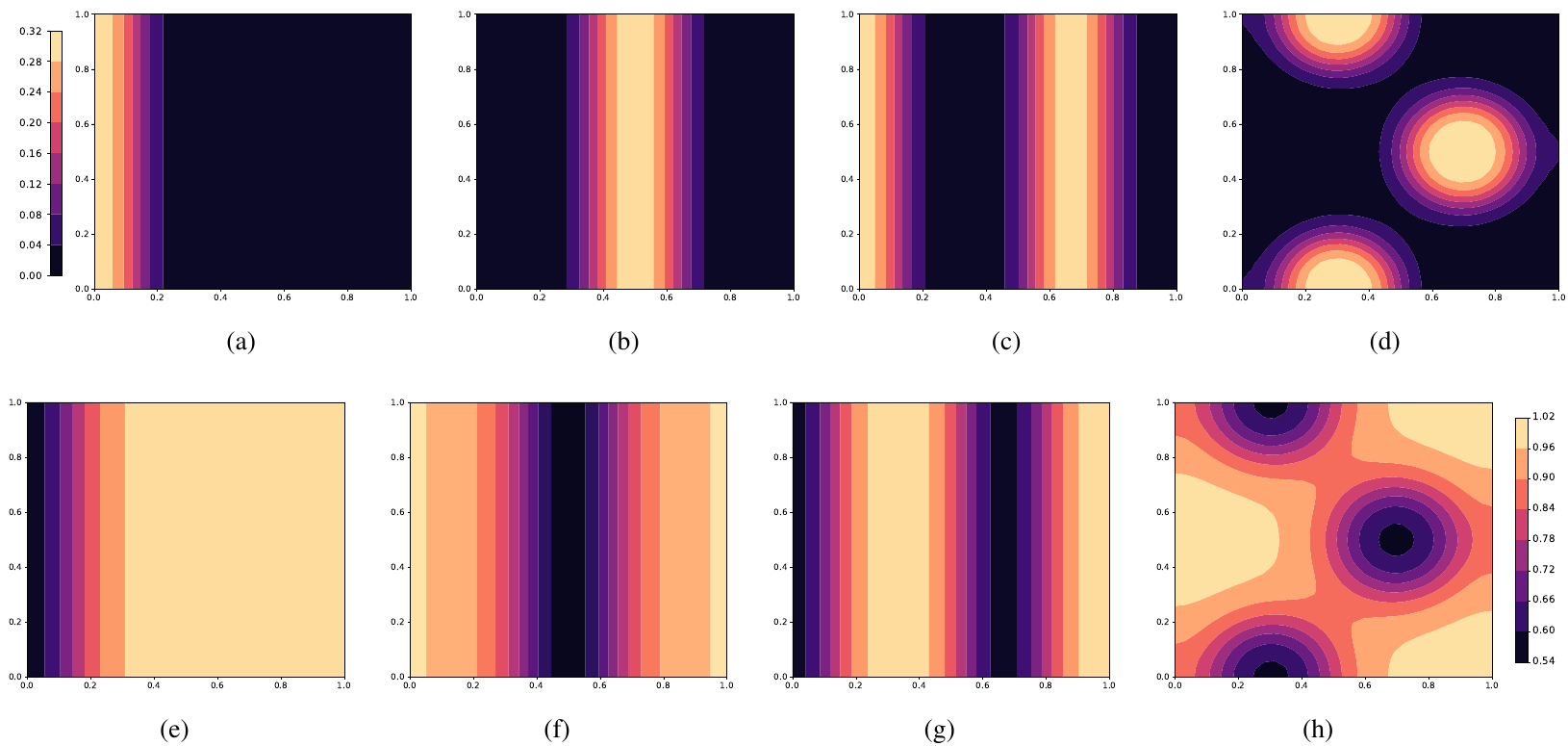}
\caption{The four known solutions of the Gray–Scott model with the chosen DE parameters to obtain observation input to HomPINN. These contour plots show the distribution of the component $A(x, y)$ (top four figures, ($(a)$, $(b)$, $(c)$, and $(d)$)) and $S(x, y)$ (bottom four figures, $(e)$, $(f)$, $(g)$, and $(h)$) on a domain of $[0, 1] \times [0, 1]$, with the horizontal and vertical axes representing x and y, respectively. The DE parameters used here: $D_A=2.5 \times 10^{-4}$, $D_S=5 \times 10^{-4}$, $\rho=4.0 \times 10^{-2}$, and $\mu=6.5\times 10^{-2}$.}\label{fig_gs_known}
\end{figure}

\begin{figure}[htbp]
\subfigure[]{
\centering
\includegraphics[width=.23\columnwidth]{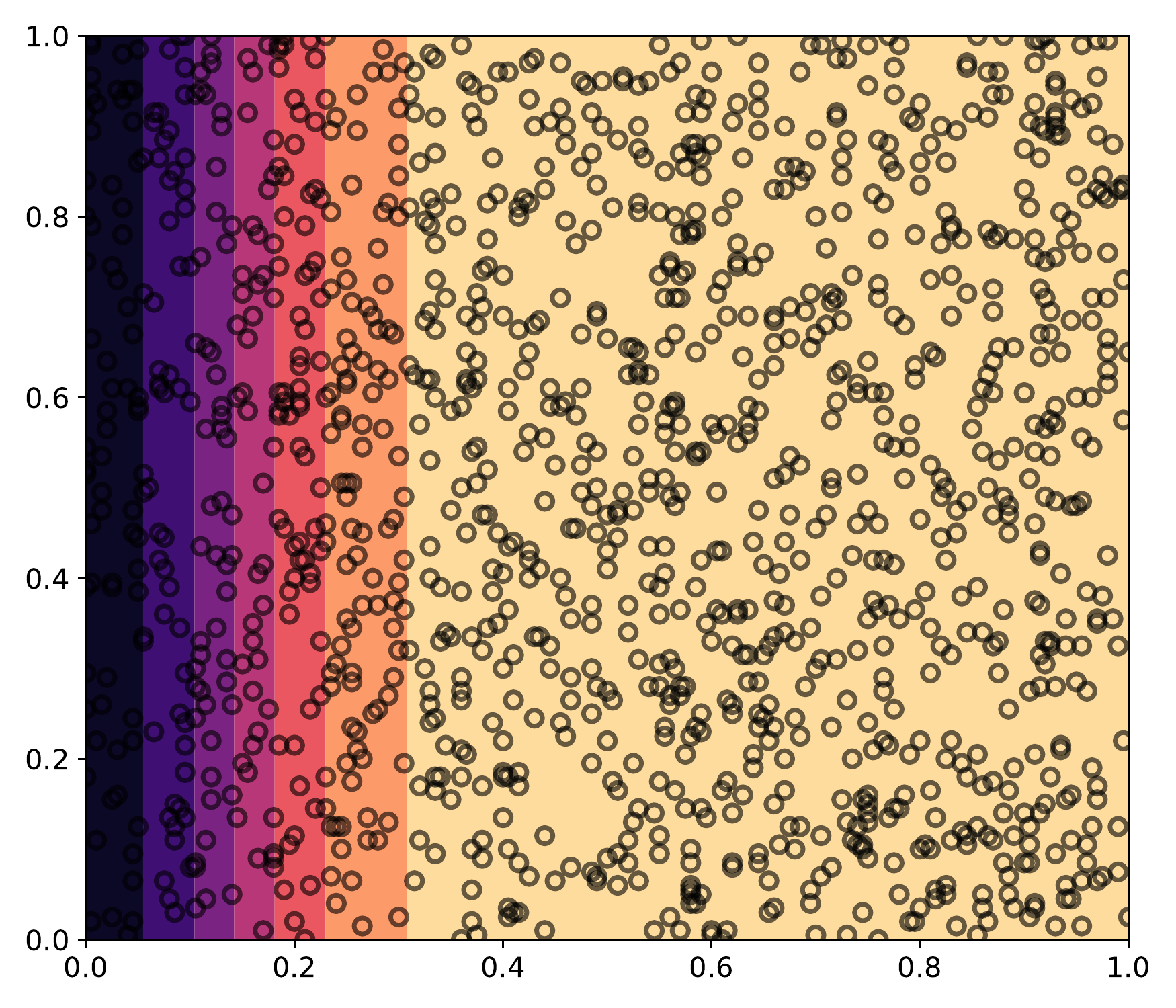}
}
\subfigure[]{
\centering
\includegraphics[width=.23\columnwidth]{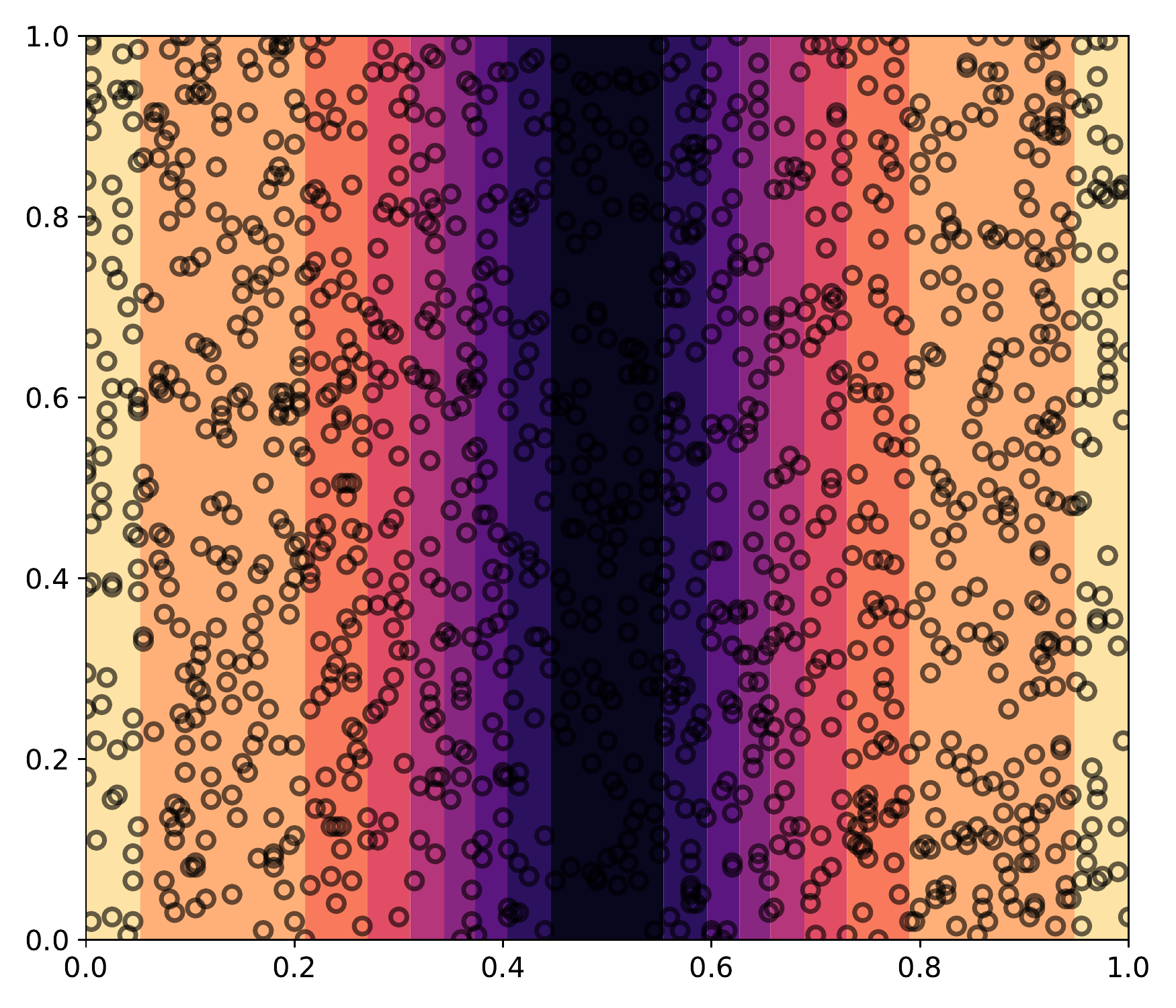}
}
\subfigure[]{
\centering
\includegraphics[width=.23\columnwidth]{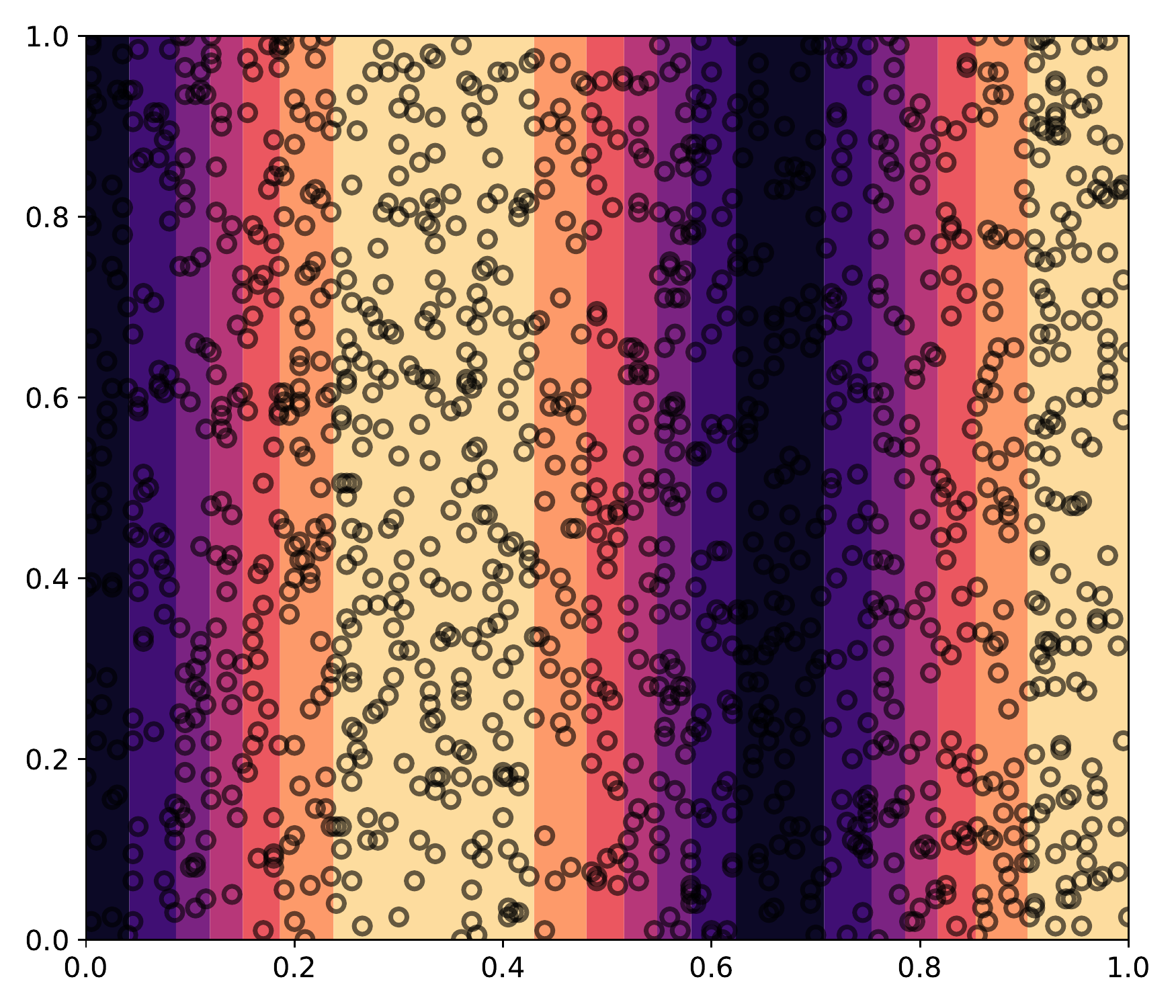}
}
\subfigure[]{
\centering
\includegraphics[width=.23\columnwidth]{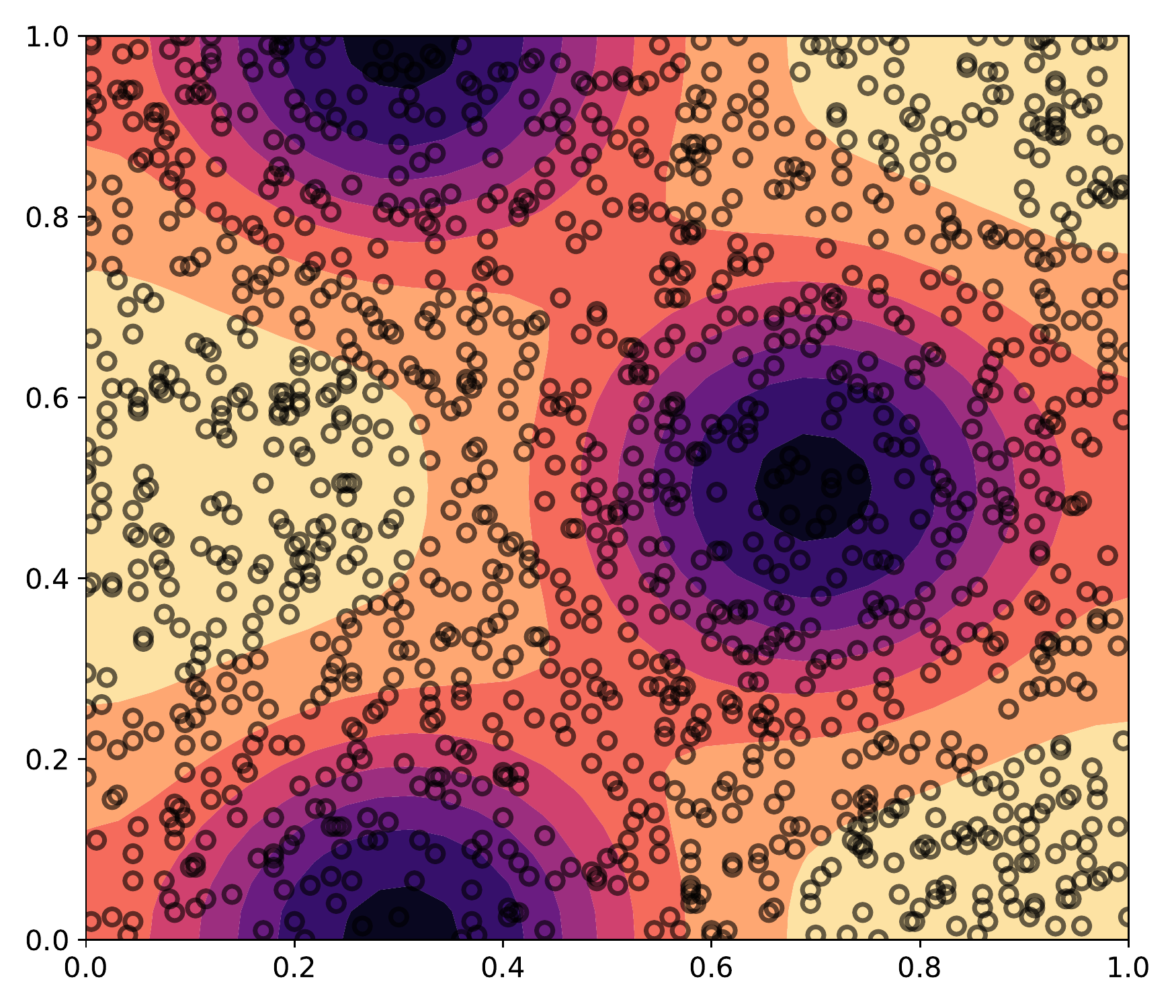}
}
\caption{Sampled observations of the Gray–Scott model in 2D. The contour plots are for the component $A(x, y)$. The horizontal and vertical axes represent x and y, respectively, and the domain is $[0, 1] \times [0, 1]$. The black circles are sampled observations.}\label{fig_gs_sample}
\end{figure}

We first randomly generate a total of 10,000 observations $\{\boldsymbol x_i,\boldsymbol u_i\}_{i=1}^{10,000}$ in the four known solutions (see solutions provided in Fig.~\ref{fig_gs_known}, and the sampled observations in Fig.~\ref{fig_gs_sample} with black circles). Then we randomly select 1,000 and 5,000 samples from $\{\boldsymbol x_i,\boldsymbol u_i\}_{i=1}^{10,000}$ to test the optimal observation selection for performing HomPINNs. 

We also investigate the network structure by changing the number of neurons in each layer (width) and the number of layers. Table \ref{table_ex3} summarizes the values of the identified parameters from HomPINNs and training loss under different conditions of the number of observations, network width, and network layers. We discover that HomPINNs can fail for several reasons, such as limited observations, inadequate network width, or insufficient network layers. HomPINNs become more effective when the number of observations exceeds 5,000, the network width is larger than 128, and the number of layers is greater than 6. When the number of given observations is too small, HomPINNs may not be able to distinguish different patterns (solutions) in the observations accurately. This can lead to overfitting of the model to the observations, which negatively impacts its generalization capacity. The number of neurons and layers in a NN determines the width and depth of the network. If one of them is too small, HomPINNs may fail to identify the unlabeled observations accurately and lead to poor generalization of potential solutions. The performance of HomPINNs can be enhanced by increasing the number of observations, network widths, and the number of network layers to identify observations and provide approximations with high accuracy.

Based on the above analysis, we opted to select 5,000 unlabeled observations randomly from the four known solutions (see Fig.~\ref{fig_gs_sample}) and configure the NN with 6 hidden layers, each of which has 128 neurons. 

\begin{figure}[htbp]
\includegraphics[width=1.\columnwidth]{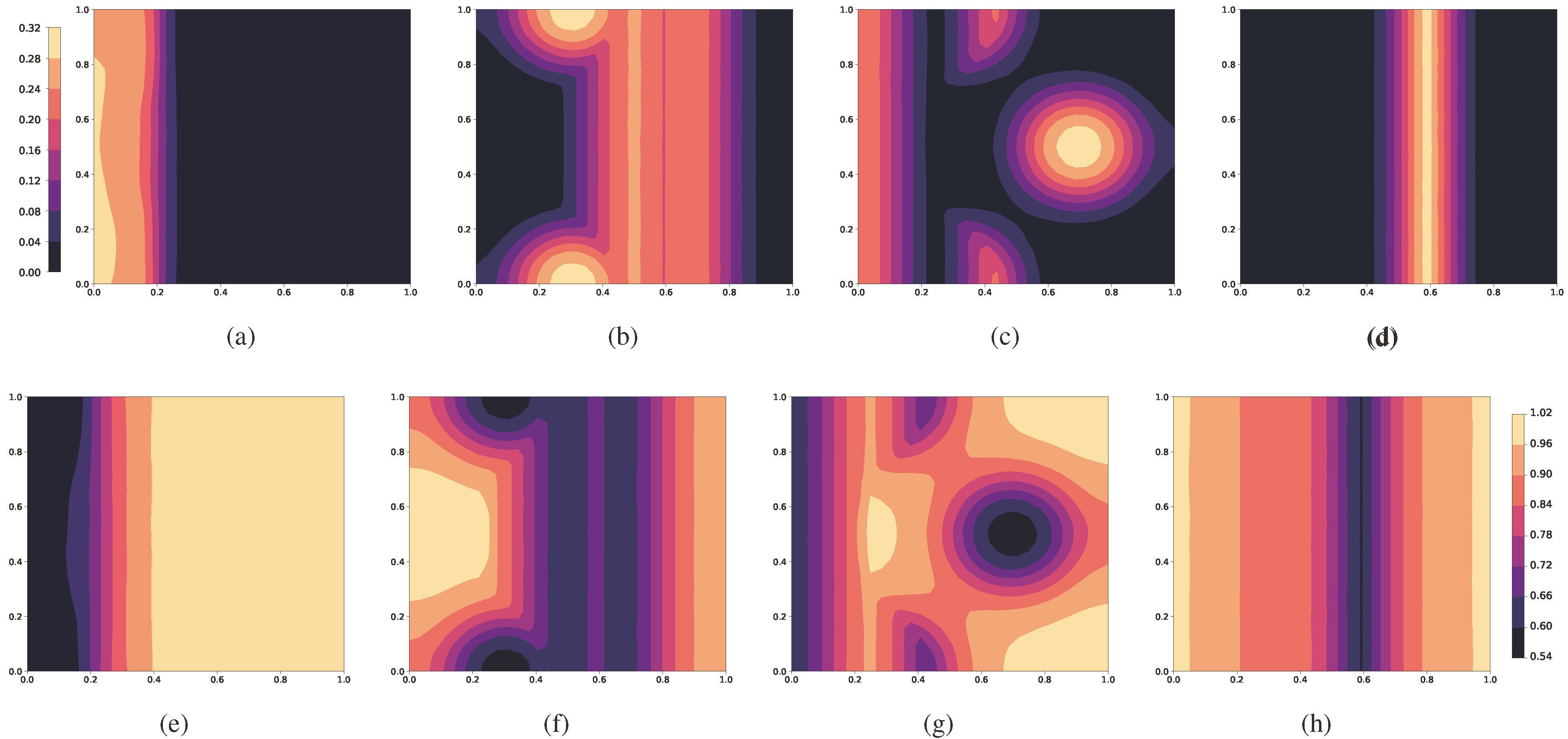}
\caption{Predictions of the Gray–Scott model after the first homotopy process. The contour plots are for components $A(x, y)$ and $S(x, y)$. The horizontal and vertical axes represent x and y, respectively. The domain is $[0, 1] \times [0, 1]$. Learned DE parameters from the first homotopy process: $D_A=2.020\times 10^{-4}$, $D_S=4.849\times 10^{-4}$, $\rho=3.678\times 10^{-2}$, and $\mu=6.882\times 10^{-2}$.}\label{fig_gs_inverse1}
\end{figure}

After the first homotopy process, the values of the identified DE parameters are $D_A=2.020\times 10^{-4}$, $D_S=4.849\times 10^{-4}$, $\rho=3.678\times 10^{-2}$, and $\mu=6.882\times 10^{-2}$ (results are given in Fig.~\ref{fig_gs_inverse1} and Table \ref{table_ex3}), which indicates that HomPINNs recover the unknown DE parameters to some extent. Nevertheless, it is unable to correctly identify different solutions associated with the observations. From Fig.~\ref{fig_gs_inverse1}, the results produced by HomPINN may mix observations from multiple solutions, yet the approximations are still satisfying the governing equation (\ref{eq_ex3_steady}).

\begin{figure}[htbp]
\includegraphics[width=1.\columnwidth]{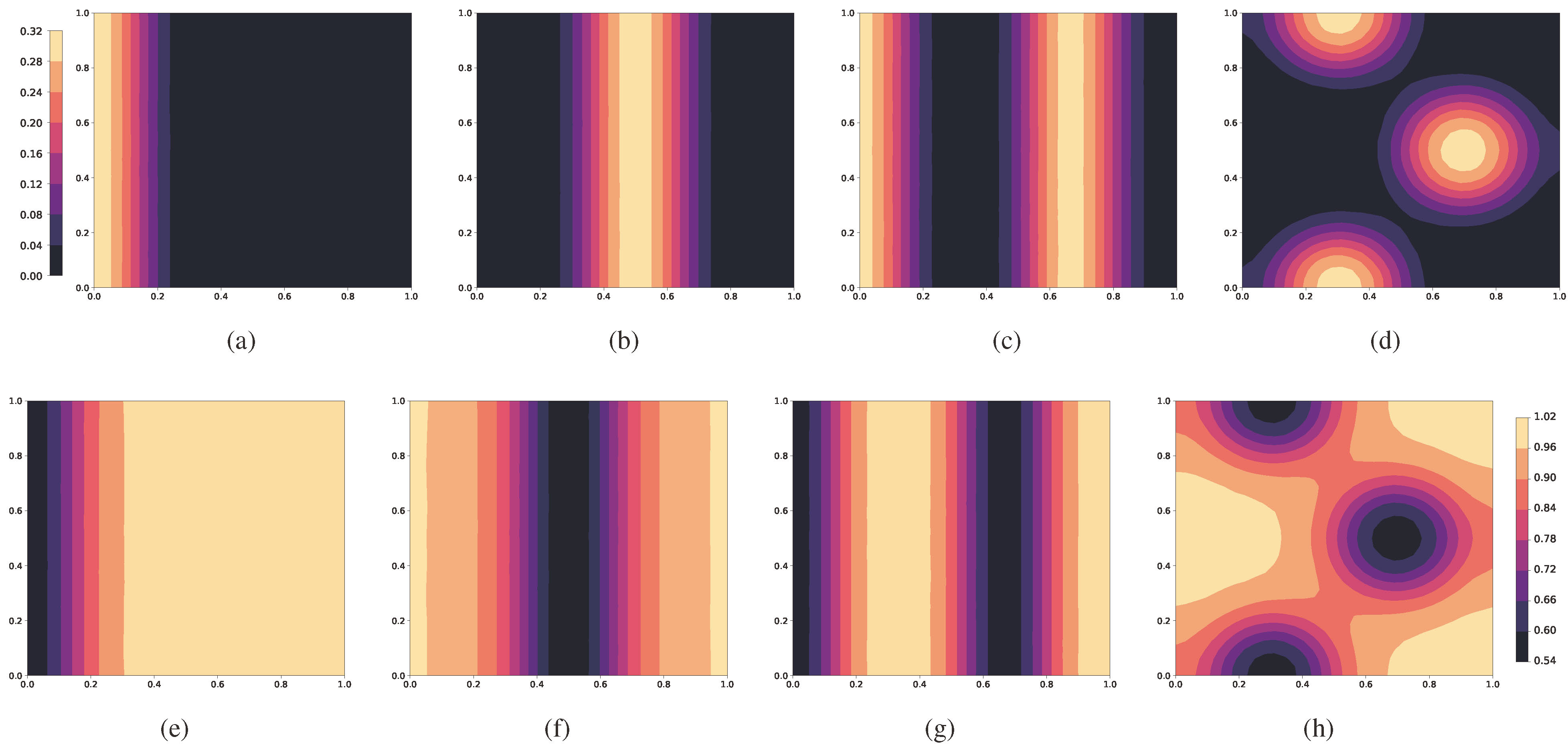}
\caption{Predictions of the Gray–Scott model from the forward HomPINNs. The contour plots of $A(x, y)$ and $S(x, y)$ are on a domain of $[0, 1] \times [0, 1]$, with the horizontal and vertical axes representing x and y, respectively. The domain is $[0, 1] \times [0, 1]$. The used DE parameters: $D_A=2.020\times 10^{-4}$, $D_S=4.849\times 10^{-4}$, $\rho=3.678\times 10^{-2}$, and $\mu=6.882\times 10^{-2}$ obtained from the first homotopy process.}\label{fig_gs_forward2}
\end{figure}

\begin{figure}[htbp]
\includegraphics[width=1.\columnwidth]{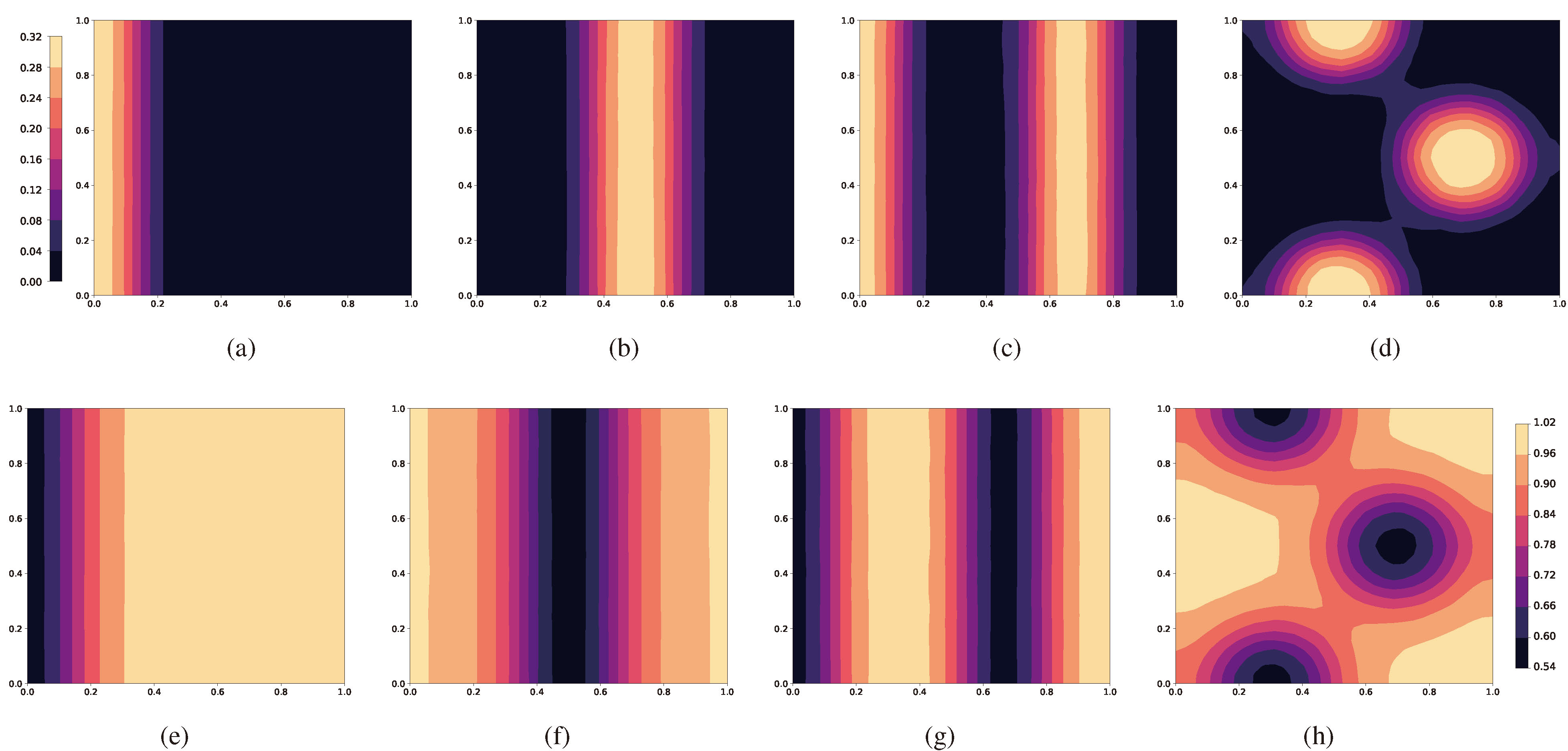}
\caption{Predictions of the Gray–Scott model after the second homotopy process. The contour plots are for components $A(x, y)$ and $S(x, y)$. The horizontal and vertical axes represent x and y, respectively. The domain is $[0, 1] \times [0, 1]$. Learned DE parameters from the second homotopy process: $D_A=2.409\times 10^{-4}$, $D_S=5.064\times 10^{-4}$, $\rho=4.076\times 10^{-2}$, and $\mu=6.432\times 10^{-2}$.}\label{fig_gs_inverse2}
\end{figure}

\begin{figure}[htbp]
\includegraphics[width=1.\columnwidth]{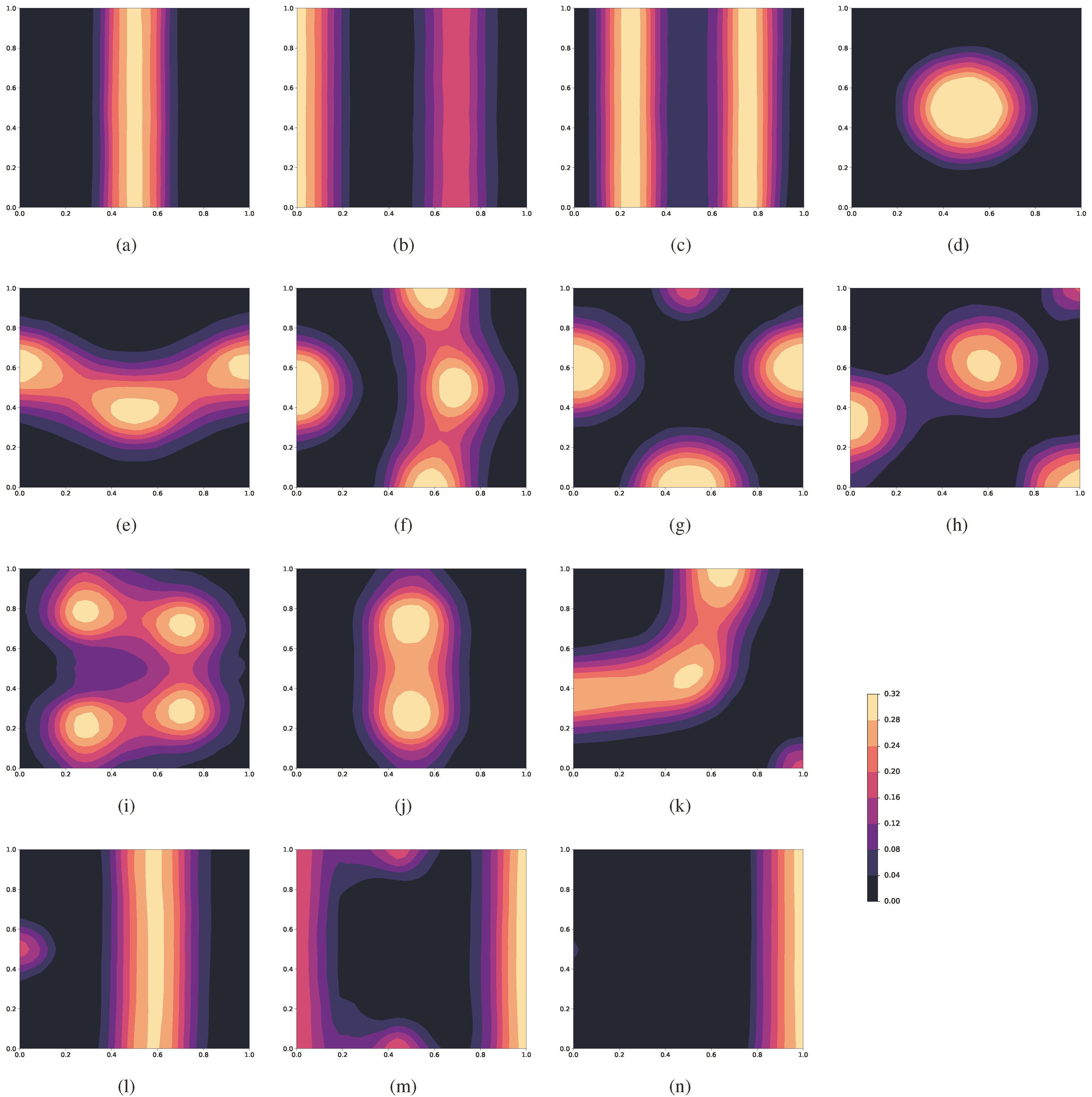}
\caption{All the solutions of (\ref{eq_ex3_steady}) from the forward HomPINNs given the parameters after the second homotopy process. The contour plots are for component $A(x, y)$. The horizontal and vertical axes represent x and y, respectively. The DE parameters used here for the forward HomPINN are: $D_A=2.409\times 10^{-4}$, $D_S=5.064\times 10^{-4}$, $\rho=4.076\times 10^{-2}$, and $\mu=6.432\times 10^{-2}$. }\label{fig_gs_unknown}
\end{figure}

We implement the strategy in Remark \ref{mark_hmsnn} to improve the accuracy. Using the parameters specified by the first homotopy process as inputs, we perform the forward HomPINNs following the process outlined in \cite{huang2022hompinns}, where the approximations correspond to $A(\cdot)$ and $S(\cdot)$ are shown in Fig.~\ref{fig_gs_forward2}. The results are closer to the correct solutions illustrated in Fig.~\ref{fig_gs_known}, although the input DE parameters from the forward HomPINNs deviate from the correct ones to some extent. Subsequently, the NN parameters in the trained forward HomPINNs along with the DE parameters specified from the first homotopy process are used to initialize the second homotopy process. With DE parameters specified in the second homotopy process being closer to the correct ones, our predictions of $A(\cdot)$ and $S(\cdot)$ are improved in the second homotopy process (see Fig.~\ref{fig_gs_inverse2}). Tables \ref{tab_gs_train1} and \ref{tab_gs_train2} provide a summary of the training outcomes for the two homotopy processes, including the DE parameters obtained in each homotopy step, the DE parameter errors, and the training losses. The results demonstrate that the DE parameters can be correctly determined by HomPINNs. Once we identify the DE parameters from HomPINN, they are input to the forward HomPINNs to discover all the solutions governed by (\ref{eq_ex3_steady}), as depicted in Fig.~\ref{fig_gs_unknown}.

% \begin{landscape}

\begin{sidewaystable}[h!]
% \begin{table}[h!]
% \scriptsize
  \centering
  \begin{threeparttable}
  \caption{Different metrics (approximate DE parameter values of HomPINNs, errors of the identified DE parameters from HomPINNs, and training loss) with different numbers of observations used to train HomPINNs, different network layers, and neurons, are recorded in the experiments. The correct DE parameters are: $D_A=2.5\times 10^{-4}$, $D_S=5.0\times 10^{-4}$, $\rho=4\times 10^{-2}$ and $\mu=6.5\times 10^{-2}$.}
  \label{table_ex3}
\begin{tabular}{cccccccccccc}
\toprule
{$N_o$}    & {\# layers} & {\# neurons} &  $D_A$ value  &  $D_S$ value  &  $\rho$ value  &  $\mu$ value & err($D_A$)  & err($D_S$)  & err($\rho$)  & err($\mu$)    & {Train loss} \\ \midrule
\multirow{12}{*}{1,000}  & \multirow{4}{*}{2}         & 32                          & 1.366E-4 & 3.417E-4 & 3.750E-2 & 6.976E-2 & 1.134E-4 & 1.583E-4 & 2.504E-3 & 4.765E-3 & 2.28E-4                    \\
                         &                            & 64                          & 1.526E-4 & 4.327E-4 & 3.808E-2 & 6.843E-2 & 9.743E-5 & 6.727E-5 & 1.924E-3 & 3.432E-3 & 6.38E-4                    \\
                         &                            & 128                         & 1.694E-4 & 4.158E-4 & 3.747E-2 & 6.914E-2 & 8.065E-5 & 8.423E-5 & 2.528E-3 & 4.141E-3 & 1.40E-4                    \\
                         &                            & 256                         & 1.756E-4 & 4.519E-4 & 3.820E-2 & 6.702E-2 & 7.442E-5 & 4.806E-5 & 1.795E-3 & 2.015E-3 & 1.71E-5                    \\ \cline{2-12} 
                         & \multirow{4}{*}{4}         & 32                          & 1.918E-4 & 4.964E-4 & 3.818E-2 & 6.797E-2 & 5.819E-5 & 3.630E-6 & 1.818E-3 & 2.971E-3 & 7.13E-5                    \\
                         &                            & 64                          & 1.872E-4 & 5.020E-4 & 3.794E-2 & 6.860E-2 & 6.283E-5 & 1.950E-6 & 2.061E-3 & 3.601E-3 & 1.02E-4                    \\
                         &                            & 128                         & 1.923E-4 & 5.222E-4 & 3.806E-2 & 6.773E-2 & 5.772E-5 & 2.215E-5 & 1.935E-3 & 2.730E-3 & 2.09E-6                    \\
                         &                            & 256                         & 1.932E-4 & 4.989E-4 & 3.758E-2 & 7.014E-2 & 5.676E-5 & 1.100E-6 & 2.422E-3 & 5.141E-3 & 2.31E-6                    \\ \cline{2-12} 
                         & \multirow{4}{*}{6}         & 32                          & 1.959E-4 & 5.409E-4 & 3.841E-2 & 6.712E-2 & 5.412E-5 & 4.085E-5 & 1.591E-3 & 2.119E-3 & 5.02E-6                    \\
                         &                            & 64                          & 2.029E-4 & 5.692E-4 & 3.809E-2 & 6.711E-2 & 4.712E-5 & 6.915E-5 & 1.908E-3 & 2.106E-3 & 1.50E-6                    \\
                         &                            & 128                         & 2.052E-4 & 6.947E-4 & 3.844E-2 & 6.578E-2 & 4.484E-5 & 1.947E-4 & 1.555E-3 & 7.800E-4 & 1.82E-6                    \\
                         &                            & 256                         & 2.045E-4 & 5.681E-4 & 3.836E-2 & 6.683E-2 & 4.555E-5 & 6.810E-5 & 1.642E-3 & 1.826E-3 & 2.86E-6     \\ \midrule
\multirow{12}{*}{5,000}  & \multirow{4}{*}{2}         & 32                          & 1.344E-4 & 3.000E-4 & 3.704E-2 & 7.195E-2 & 1.156E-4 & 2.000E-4 & 2.964E-3 & 6.949E-3 & 1.28E-3                    \\
                         &                            & 64                          & 1.675E-4 & 4.318E-4 & 3.799E-2 & 6.742E-2 & 8.248E-5 & 6.821E-5 & 2.010E-3 & 2.418E-3 & 1.00E-3                    \\
                         &                            & 128                         & 1.949E-4 & 4.613E-4 & 3.801E-2 & 6.673E-2 & 5.508E-5 & 3.866E-5 & 1.987E-3 & 1.729E-3 & 2.79E-4                    \\
                         &                            & 256                         & 1.902E-4 & 4.574E-4 & 3.791E-2 & 6.770E-2 & 5.979E-5 & 4.256E-5 & 2.090E-3 & 2.704E-3 & 5.19E-4                    \\ \cline{2-12} 
                         & \multirow{4}{*}{4}         & 32                          & 1.923E-4 & 4.895E-4 & 3.729E-2 & 6.984E-2 & 5.775E-5 & 1.046E-5 & 2.708E-3 & 4.836E-3 & 1.28E-4                    \\
                         &                            & 64                          & 1.907E-4 & 4.854E-4 & 3.731E-2 & 7.006E-2 & 5.928E-5 & 1.460E-5 & 2.689E-3 & 5.064E-3 & 1.98E-5                    \\
                         &                            & 128                         & 1.968E-4 & 4.812E-4 & 3.684E-2 & 7.120E-2 & 5.317E-5 & 1.885E-5 & 3.156E-3 & 6.201E-3 & 8.95E-6                    \\
                         &                            & 256                         & 1.984E-4 & 4.837E-4 & 3.694E-2 & 7.036E-2 & 5.161E-5 & 1.626E-5 & 3.064E-3 & 5.356E-3 & 4.96E-6                    \\ \cline{2-12} 
                         & \multirow{4}{*}{6}         & 32                          & 1.980E-4 & 4.866E-4 & 3.684E-2 & 6.800E-2 & 5.199E-5 & 1.337E-5 & 3.161E-3 & 2.997E-3 & 6.60E-6                    \\
                         &                            & 64                          & 2.001E-4 & 4.852E-4 & 3.682E-2 & 6.811E-2 & 4.991E-5 & 1.476E-5 & 3.179E-3 & 3.113E-3 & 5.25E-6                    \\
                         &                            & 128                         & 2.020E-4 & 4.849E-4 & 3.678E-2 & 6.882E-2 & 4.798E-5 & 1.507E-5 & 3.224E-3 & 3.816E-3 & 6.38E-7                    \\
                         &                            & 256                         & 1.893E-4 & 4.837E-4 & 3.668E-2 & 7.606E-2 & 6.069E-5 & 1.628E-5 & 3.317E-3 & 1.106E-2 & 1.20E-6                    \\ \midrule
\multirow{12}{*}{10,000} & \multirow{4}{*}{2}         & 32                          & 1.785E-4 & 4.607E-4 & 3.797E-2 & 6.692E-2 & 7.154E-5 & 3.926E-5 & 2.032E-3 & 1.924E-3 & 1.98E-3                    \\
                         &                            & 64                          & 1.778E-4 & 4.401E-4 & 3.791E-2 & 6.674E-2 & 7.217E-5 & 5.987E-5 & 2.087E-3 & 1.742E-3 & 6.49E-4                    \\
                         &                            & 128                         & 1.949E-4 & 4.656E-4 & 3.766E-2 & 6.772E-2 & 5.514E-5 & 3.444E-5 & 2.337E-3 & 2.724E-3 & 1.05E-3                    \\
                         &                            & 256                         & 1.947E-4 & 4.668E-4 & 3.765E-2 & 6.765E-2 & 5.532E-5 & 3.316E-5 & 2.351E-3 & 2.646E-3 & 5.19E-4                    \\ \cline{2-12} 
                         & \multirow{4}{*}{4}         & 32                          & 2.021E-4 & 4.897E-4 & 3.721E-2 & 6.917E-2 & 4.787E-5 & 1.032E-5 & 2.789E-3 & 4.167E-3 & 5.16E-5                    \\
                         &                            & 64                          & 2.012E-4 & 4.851E-4 & 3.677E-2 & 6.865E-2 & 4.883E-5 & 1.487E-5 & 3.233E-3 & 3.647E-3 & 2.23E-5                    \\
                         &                            & 128                         & 2.064E-4 & 4.829E-4 & 3.688E-2 & 6.859E-2 & 4.364E-5 & 1.711E-5 & 3.119E-3 & 3.595E-3 & 1.13E-5                    \\
                         &                            & 256                         & 2.067E-4 & 4.844E-4 & 3.686E-2 & 6.880E-2 & 4.329E-5 & 1.565E-5 & 3.137E-3 & 3.796E-3 & 4.11E-6                    \\ \cline{2-12} 
                         & \multirow{4}{*}{6}         & 32                          & 2.029E-4 & 4.819E-4 & 3.689E-2 & 6.771E-2 & 4.710E-5 & 1.815E-5 & 3.106E-3 & 2.711E-3 & 8.29E-6                    \\
                         &                            & 64                          & 2.053E-4 & 4.855E-4 & 3.680E-2 & 6.789E-2 & 4.471E-5 & 1.455E-5 & 3.199E-3 & 2.892E-3 & 2.61E-6                    \\
                         &                            & 128                         & 2.077E-4 & 4.889E-4 & 3.683E-2 & 6.854E-2 & 4.230E-5 & 1.108E-5 & 3.168E-3 & 3.536E-3 & 3.75E-7                    \\
                         &                            & 256                         & 2.086E-4 & 4.925E-4 & 3.702E-2 & 6.874E-2 & 4.143E-5 & 7.540E-6 & 2.982E-3 & 3.744E-3 & 1.61E-6                   \\ \bottomrule
\end{tabular}
    \end{threeparttable}
% \end{table}
\end{sidewaystable}

% \end{landscape}

% \clearpage

\begin{table}[htbp!]\centering
\caption{Approximate values and errors of the identified DE parameters from the first homotopy process, and training losses among the first homotopy process. The correct DE parameters are $D_A = 2.5 \times 10^{-4}$, $D_S = 5 \times 10^{-4}$, $\rho = 0.04$, and $\mu = 0.065$.}\label{tab_gs_train1}
\begin{tabular}{ccccccccccc}
\toprule
{$k$} & $D_A$ value  &  $D_S$ value  &  $\rho$ value  &  $\mu$ value & & err($D_A$)  & err($D_S$)  & err($\rho$)  & err($\mu$)    & {Train loss} \\ \midrule
                     & 1.000E-5 & 2.000E-5 & 1.000E-3 & 1.000E-3 &  & 2.400E-4 & 4.800E-4 & 3.900E-2 & 6.400E-2 & 6.157E-3                   \\
1                    & 1.620E-4 & 4.709E-4 & 3.820E-2 & 8.931E-2 &  & 8.800E-5 & 2.911E-5 & 1.800E-3 & 2.431E-2 & 5.051E-4                   \\
2                    & 1.875E-4 & 4.816E-4 & 3.710E-2 & 7.155E-2 &  & 6.252E-5 & 1.840E-5 & 2.904E-3 & 6.552E-3 & 2.296E-5                   \\
3                    & 1.948E-4 & 4.821E-4 & 3.688E-2 & 6.954E-2 &  & 5.521E-5 & 1.795E-5 & 3.121E-3 & 4.537E-3 & 2.643E-5                   \\
4                    & 1.970E-4 & 4.847E-4 & 3.688E-2 & 6.909E-2 &  & 5.297E-5 & 1.526E-5 & 3.120E-3 & 4.088E-3 & 7.933E-6                   \\
5                    & 1.989E-4 & 4.836E-4 & 3.678E-2 & 6.897E-2 &  & 5.110E-5 & 1.639E-5 & 3.221E-3 & 3.965E-3 & 4.712E-6                   \\
6                    & 1.999E-4 & 4.840E-4 & 3.675E-2 & 6.884E-2 &  & 5.010E-5 & 1.602E-5 & 3.248E-3 & 3.842E-3 & 2.091E-6                   \\
7                    & 2.008E-4 & 4.841E-4 & 3.677E-2 & 6.884E-2 &  & 4.917E-5 & 1.586E-5 & 3.234E-3 & 3.835E-3 & 2.057E-6                   \\
8                    & 2.013E-4 & 4.842E-4 & 3.676E-2 & 6.885E-2 &  & 4.872E-5 & 1.579E-5 & 3.236E-3 & 3.848E-3 & 1.307E-6                   \\
9                    & 2.016E-4 & 4.848E-4 & 3.677E-2 & 6.883E-2 &  & 4.837E-5 & 1.522E-5 & 3.230E-3 & 3.829E-3 & 7.243E-7                   \\
10                   & 2.020E-4 & 4.849E-4 & 3.678E-2 & 6.882E-2 &  & 4.798E-5 & 1.507E-5 & 3.224E-3 & 3.816E-3 & 6.376E-7  \\ \bottomrule        
\end{tabular}
\end{table}

\begin{table}[htbp!]\centering
\caption{Approximate values and errors of the identified DE parameters from the second homotopy process, and training losses among the second homotopy process. The correct DE parameter are $D_A = 2.5 \times 10^{-4}$, $D_S = 5 \times 10^{-4}$, $\rho = 0.04$, and $\mu = 0.065$.}\label{tab_gs_train2}
\begin{tabular}{cccccccccccc}
\toprule
{$k$} & $D_A$ value  &  $D_S$ value  &  $\rho$ value  &  $\mu$ value & & err($D_A$)  & err($D_S$)  & err($\rho$)  & err($\mu$)    & {Train loss} \\ \midrule
    & 2.020E-4 & 4.849E-4 & 3.678E-2 & 6.882E-2 &  & 4.798E-5 & 1.507E-5 & 3.224E-3 & 3.816E-3 & 1.339E-5   \\
1   & 2.384E-4 & 5.057E-4 & 4.072E-2 & 6.424E-2 & & 1.159E-5 & 5.699E-6 & 7.212E-4 & 7.581E-4 & 9.054E-6    \\
2   & 2.397E-4 & 5.059E-4 & 4.073E-2 & 6.428E-2 & & 1.028E-5 & 5.889E-6 & 7.286E-4 & 7.199E-4 & 5.097E-6    \\
3   & 2.404E-4 & 5.061E-4 & 4.074E-2 & 6.429E-2 & & 9.638E-6 & 6.098E-6 & 7.407E-4 & 7.072E-4 & 3.194E-6    \\
4   & 2.406E-4 & 5.063E-4 & 4.075E-2 & 6.430E-2 & & 9.430E-6 & 6.298E-6 & 7.539E-4 & 7.009E-4 & 1.891E-6    \\
5   & 2.407E-4 & 5.064E-4 & 4.076E-2 & 6.430E-2 & & 9.268E-6 & 6.360E-6 & 7.585E-4 & 7.009E-4 & 1.129E-6    \\
6   & 2.408E-4 & 5.063E-4 & 4.076E-2 & 6.430E-2 & & 9.222E-6 & 6.339E-6 & 7.610E-4 & 7.009E-4 & 7.102E-7    \\
7   & 2.409E-4 & 5.064E-4 & 4.076E-2 & 6.431E-2 & & 9.129E-6 & 6.360E-6 & 7.589E-4 & 6.945E-4 & 4.132E-7    \\
8   & 2.409E-4 & 5.064E-4 & 4.076E-2 & 6.431E-2 & & 9.106E-6 & 6.396E-6 & 7.610E-4 & 6.945E-4 & 2.435E-7    \\
9   & 2.409E-4 & 5.063E-4 & 4.076E-2 & 6.432E-2 & & 9.083E-6 & 6.350E-6 & 7.585E-4 & 6.818E-4 & 1.642E-7    \\
10  & 2.409E-4 & 5.064E-4 & 4.076E-2 & 6.432E-2 & & 9.106E-6 & 6.360E-6 & 7.585E-4 & 6.818E-4 & 1.249E-7      \\ \bottomrule        
\end{tabular}
\end{table}

\section{Conclusion and Discussion}\label{sec_discuss}

To tackle the complex behavior arising from non-uniqueness, symmetry, and bifurcations in the solution space, we proposed a novel HomPINN method for solving inverse problems in nonlinear DEs with multiple solutions. The proposed method utilizes a combination of homotopy continuation and NNs in the training process to estimate unknown parameters in the DEs and is able to produce accurate predictions. We evaluated the effectiveness of our proposed method by conducting experiments on different tasks. For DEs in one dimension, we explore the effect of various factors on the model's performance, such as {diverse DE parameter initialization}, estimates of the number of solutions, the number of observations, and observations that include partial DE solutions. Next, the two-dimensional Gray-Scott problem is investigated to identify correct DE parameters and recover all potential solutions. Through the tests conducted, we first determine the optimal number of observations and network structure. With the determined hyper-parameters, HomPINN successfully specifies the values of the four parameters in the Gray-Scott problem, although the input observations only include four solutions. All the solutions to the Grey-Scott problem are finally recovered with HomPINN, through the use of the DE parameters identified by HomPINNs.

Our experimental results indicate the following advantages of the proposed model: 1. HomPINN offers a flexible and adaptable solution for inverse problems; 2. {Compared with PINNs, HomPINNs are superior in identifying unlabeled observations from multiple solutions.} 3. HomPINN is able to tackle both one and two-dimensional problems; 4. the optimal number of observations and network structure has been identified in our tests; 5. HomPINN is able to solve inverse problems even if we do not know all the solutions of a DE; 5. It is possible to discover other solutions of a DE that are not available in observations.

In the future, our primary objective will be to refine our techniques, starting with the development of methods for the automatic identification of hyper-parameter selections. {Building on this foundation, we will then extend our approach to addressing the impact of noisy training data, enhancing its applicability in real-world scenarios.} Finally, to further ascertain the versatility of our methodology, we intend to test it on a broader range of problems, especially those characterized by complex geometries and boundary conditions.

\section{Acknowledgements}
W.H. was supported by the National Science Foundation award DMS-2052685 and the National Institutes of Health award 1R35GM146894. G.L. and H.Z. gratefully acknowledge the support of the National Science Foundation (DMS-2053746, DMS-2134209, ECCS-2328241, and OAC-2311848), and U.S. Department of Energy (DOE) Office of Science Advanced Scientific Computing Research program DE-SC0021142, DE-SC0023161, and the Uncertainty Quantification for Multifidelity Operator Learning (MOLUcQ) project (Project No. 81739), and DOE – Fusion Energy Science,  under grant number: DE-SC0024583. 
\clearpage

\bibliographystyle{ref-style}
\bibliography{refs}

\end{document}